\documentclass[10pt,twocolumn,letterpaper]{article}

\usepackage{iccv}
\usepackage{times}
\usepackage{epsfig}
\usepackage{graphicx}
\usepackage{subfigure}
\usepackage{amsmath}
\usepackage{amssymb}
\usepackage{soul}
\usepackage{algorithm}
\usepackage{algorithmic}
\usepackage{enumitem}
\usepackage{wrapfig}
\usepackage{array}

\DeclareMathOperator*{\argmin}{arg\,min}

\newcommand{\bx}{\mathbf{x}}
\newcommand{\bc}{\mathbf{c}}

\newcommand{\bA}{\mathbf{A}}
\newcommand{\cX}{\mathcal{X}}

\newcommand{\bW}{\mathbf{W}}

\newcommand{\bH}{\mathbf{H}}
\newcommand{\bh}{\mathbf{h}}
\newcommand{\bm}{\mathbf{m}}
\newcommand{\bM}{\mathbf{M}}
\newcommand{\bS}{\mathbf{S}}

\newcommand{\bF}{\mathbf{F}}

\usepackage[pagebackref=true,breaklinks=true,letterpaper=true,colorlinks,bookmarks=false]{hyperref}

\iccvfinalcopy 

\def\iccvPaperID{921} 

\ificcvfinal\fi

\title{Correspondence Insertion for As-Projective-As-Possible Image Stitching}

\author{William X. Liu, Tat-Jun Chin\\
School of Computer Science\\The University of Adelaide\\
}
\begin{document}

\maketitle

\begin{abstract}
Spatially varying warps are increasingly popular for image alignment. In particular, as-projective-as-possible (APAP) warps have been proven effective for accurate panoramic stitching, especially in cases with significant depth parallax that defeat standard homographic warps. However, estimating spatially varying warps requires a sufficient number of feature matches. In image regions where feature detection or matching fail, the warp loses guidance and is unable to accurately model the true underlying warp, thus resulting in poor registration. In this paper, we propose a correspondence insertion method for APAP warps, with a focus on panoramic stitching. Our method automatically identifies misaligned regions, and inserts appropriate point correspondences to increase the flexibility of the warp and improve alignment. Unlike other warp varieties, the underlying projective regularisation of APAP warps reduces overfitting and geometric distortion, despite increases to the warp complexity. Comparisons with recent techniques for parallax-tolerant image stitching demonstrate the effectiveness and simplicity of our approach.
\end{abstract}

\section{Introduction}

\vspace{-0.5em}

\begin{figure}
\centering
\subfigure[Input images with verified keypoint correspondences.\label{fig:gh-full}]{\includegraphics[width = 0.9\linewidth,height=0.1\textheight]{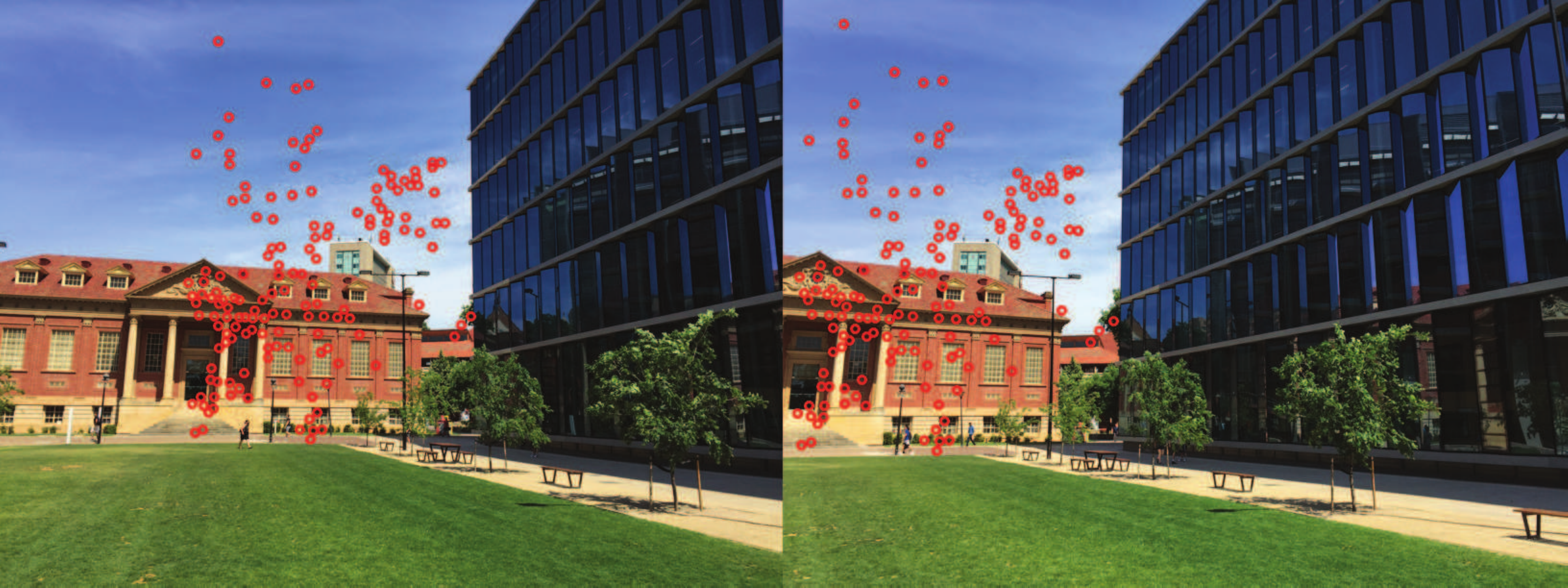}}\\
\vspace{-0.5em}
\subfigure[Image stitching result using APAP warp.\label{fig:mdlt-crop}]{\includegraphics[width = 0.9\linewidth,height=0.1\textheight]{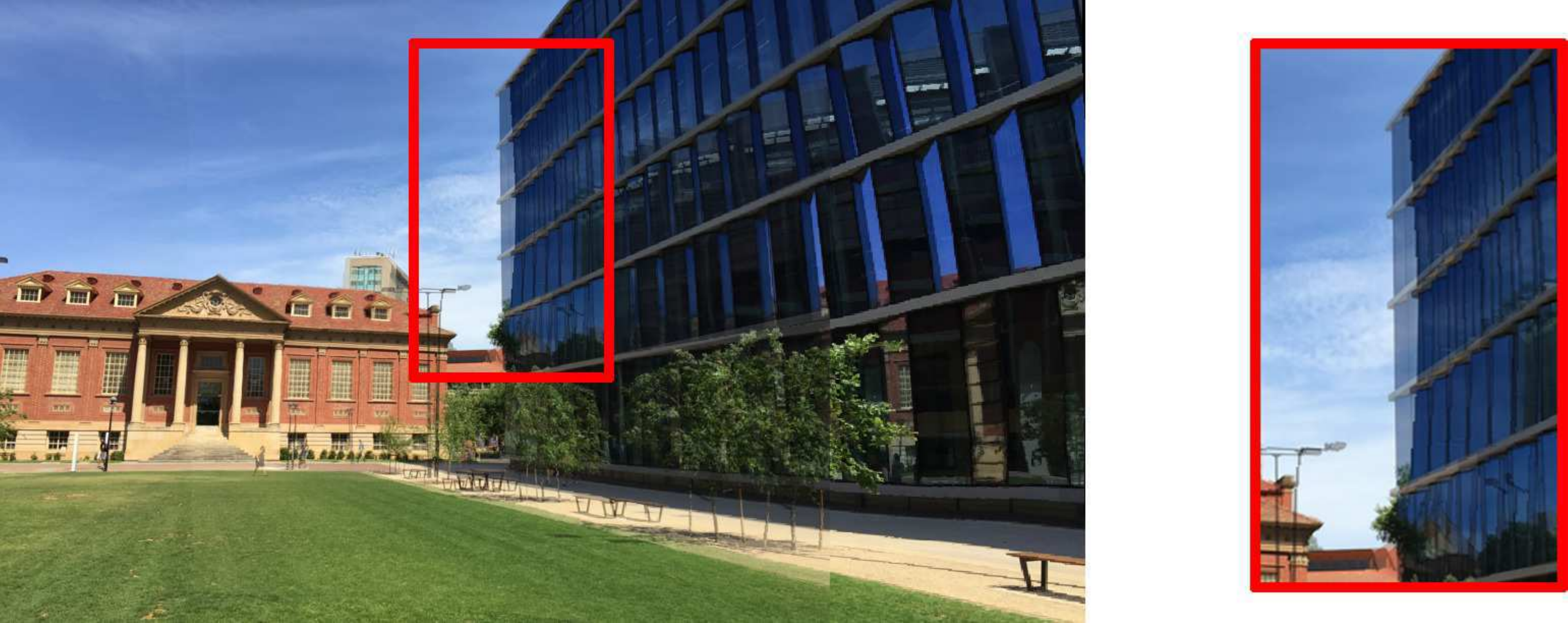}}\\
\vspace{-0.5em}
\subfigure[Result after automatically optimizing $25$ new correspondences (indicated as yellow crosses) using our method.\label{fig:ci-crop}]{\includegraphics[width = 0.9\linewidth,height=0.1\textheight]{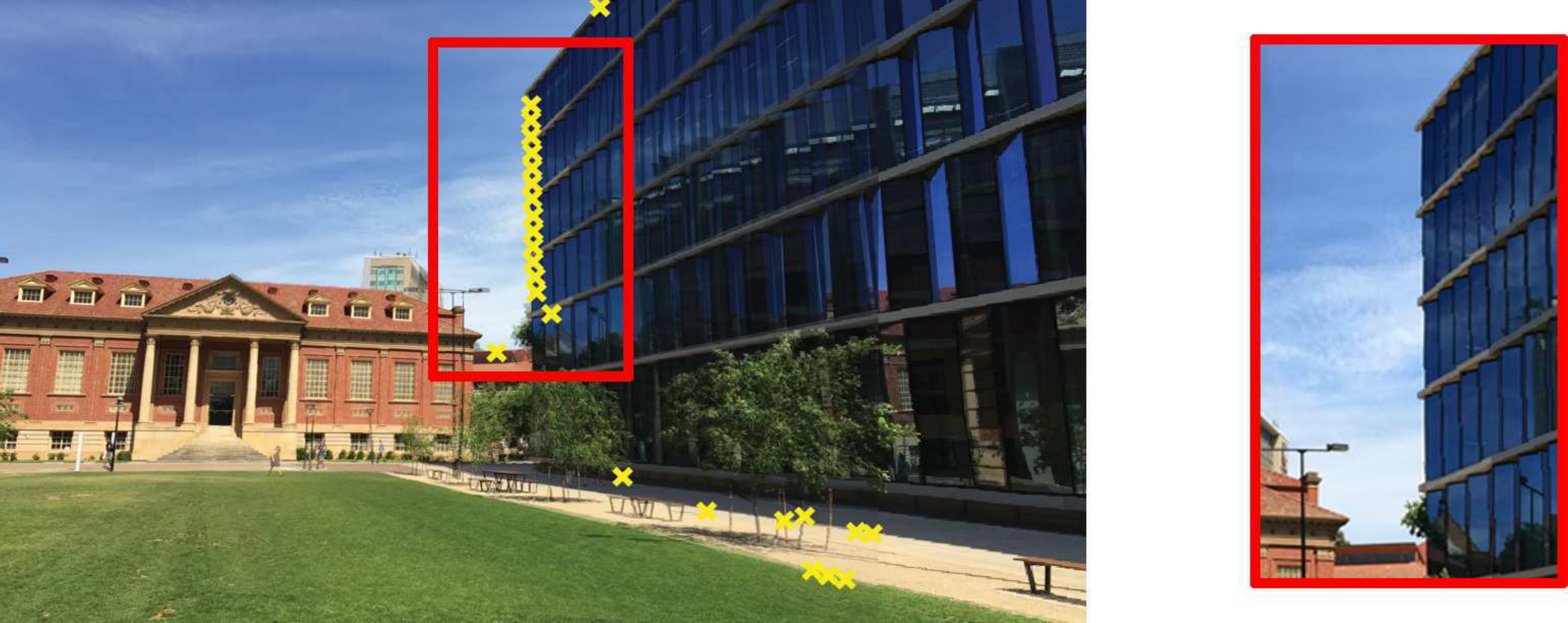}}\\
\label{fig:overview}
\caption{(a) Two input images with verified keypoint correspondences; (b) Although APAP warp is spatially varying, without sufficient keypoint correspondences, the flexibility of the warp cannot be realised and the overlap area cannot be aligned well; (c) Our proposed correspondence insertion algorithm automatically inserts and optimizes new correspondences to improve the alignment.\vspace{-1em}}
\end{figure}

The standard pipeline used in panoramic image stitching~\cite{szeliski04,brown07} begins by detecting and matching local features or keypoints across the input images. A robust technique such as RANSAC is invoked to estimate the alignment functions, usually projective transformations (i.e. homographies), based on the feature matches. Bundle adjustment is then conducted to refine the homographies, before blending and compositing of the overlap pixels take place.

The usage of homographic warps for image stitching has been questioned~\cite{gao11,zaragoza13}, since it carries the assumptions that the images were taken under pure rotational motions, or that the scene is sufficiently far away such that it is effectively planar - conditions unlikely to be satisfied in casual photography. As a result, misalignment effects or ``ghosting" inevitably occur, and relatively costly postprocessing routines are necessary to rectify or conceal the errors.

Spatially varying warps have been proposed as alternatives to homographic warps~\cite{liu09,gao11,zaragoza13,chang14}. Such warps can better account for the effects of parallax when aligning the overlap regions. In particular, as-projective-as-possible (APAP) warps~\cite{zaragoza13} interpolate the data flexibly, while maintaining a global projective trend so as to extrapolate correctly. Half-projective warps~\cite{chang14} improve upon APAP by preventing excessive stretching when extrapolating.

Ultimately, spatially varying warps are only as flexible as warranted by available feature matches. Without a sufficiently dense sampling of the underlying interpolant, the warp reduces to the baseline warp (similarity~\cite{liu09}, projective~\cite{zaragoza13}), thus defeating its spatially varying ability. A large number of feature matches are thus required to obtain good alignment, especially in areas with parallax where the true alignment function deviates from a simple homography. There is no guarantee, however, that feature matches are produced uniformly in the overlap area; see Fig.~\ref{fig:overview}.

A potential remedy is to use flow-based methods to obtain dense flow fields (e.g.,~\cite{brox11,liuce09,liuce11}). However, these methods are geared towards feature tracking in videos for motion analysis and segmentation problems. Our tests show that flow-based methods often fail in producing accurate correspondences for wide-baseline image stitching, especially in scenes with repetitive textures; see Fig.~\ref{fig:warp_flowbased}.

On the other hand, direct methods~\cite{szeliski97,shum00} align images based on pixel intensities, without requiring \emph{a priori} established feature matches. Thus the problem of undersampling the true warp does not exist. Data-driven approaches to adaptively increase the warp complexity (e.g., by adding centers or control points in splines~\cite{cham99,marsland03,bartoli04}) are also available. It is well known, however, that direct methods easily get stuck in local minima, thus necessitating coarse-to-fine registration strategies which can be inefficient~\cite{baker04}.

\begin{figure}[h]
\centering
\subfigure[Result with simple linear blending (pixel averaging).]
{\includegraphics[width = 0.99\columnwidth]
{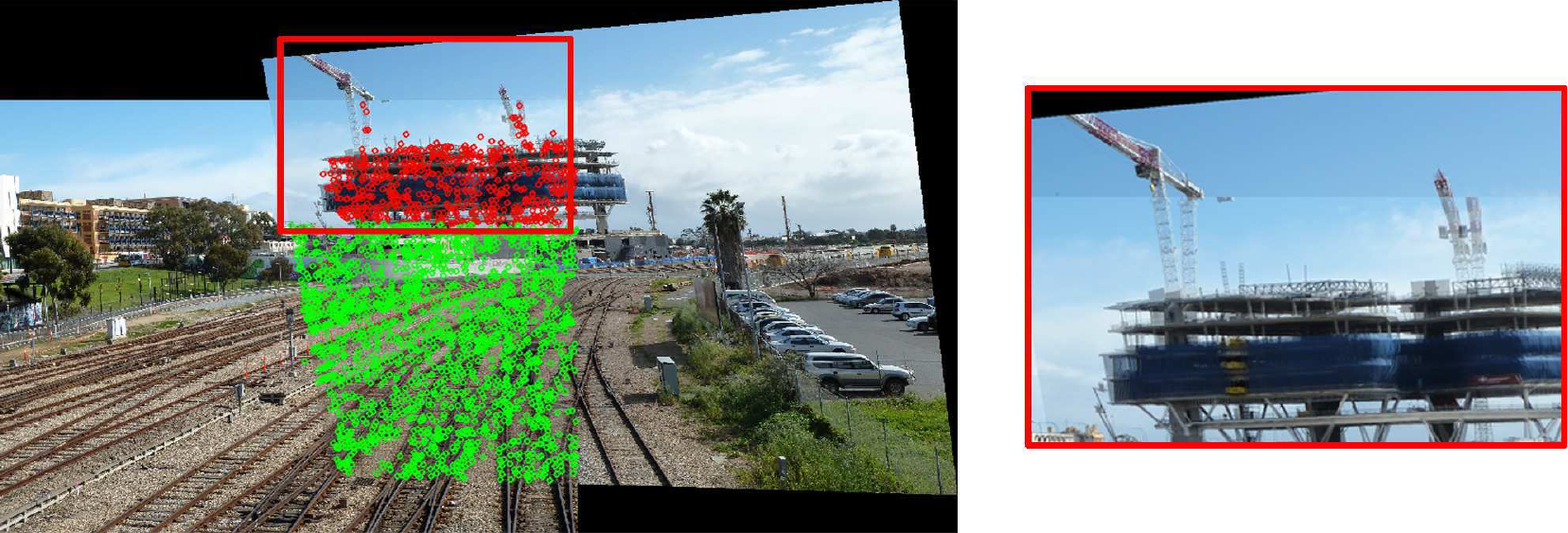}}
\subfigure[Result after seam cut pixel selection to remove ghosting.]
{\includegraphics[width = 0.99\columnwidth]
{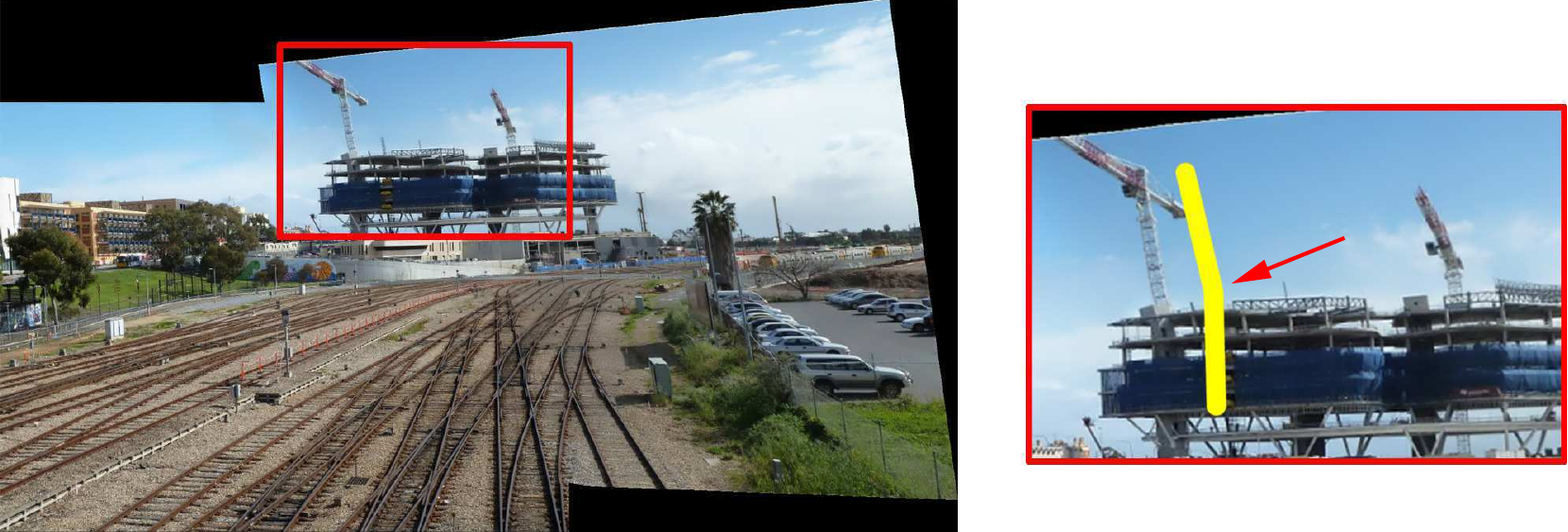}}
\caption{(a) Parallax-tolerant image stitching finds a homography that aligns a local region as well as possible. Here, green points are correspondences that are fitted by the local homography. Expectedly, regions that do not lie on the same plane cannot be aligned well; (b) Seam cut removes ghosting, but produces perceptually awkward results; note that the left crane appears to be bent. This result was taken directly from the project website of~{\cite{zhang14}}.}
\label{fig:liufeng}
\end{figure}

An alternative idea is that perfect alignment throughout the overlap region is unnecessary~{\cite{zhang14}}. Rather, images need only be aligned well in a local area, and a randomized algorithm was proposed to find a local homography. Seam cut~{\cite{agarwala04}} is then used to remove misalignments elsewhere. Such an approach is heavily dependent on \emph{postprocessing} by seam cut. However, if misalignments are too severe, seam cut may not produce geometrically correct results~{\cite{jia08}}. This will occur when the true alignment function deviates significantly from a homography, e.g., when there are two apparent planes; see {Fig.~\ref{fig:liufeng}}. More crucially, this method is reliant on the existing set of keypoint matches and cannot introduce new correspondences.

\vspace{-1em}

\paragraph{Contributions}

Our work differs from~\cite{zhang14} in that we attempt to accurately align the images \emph{throughout} the overlap area before compositing. Specifically, in correspondence-poor regions, we propose a \emph{correspondence insertion} algorithm such that a good warping function can still be estimated. We show how correspondence search can be accomplished for moving direct linear transformation (MDLT), which is the estimation method for APAP warps~\cite{zaragoza13}. We also highlight the simplicity of our data-driven warp adaptation scheme over previous spline-based center insertion techniques~\cite{cham99,marsland03,bartoli04}. On panoramic mosaicing problems that are challenging, we show that our approach achieves accurate alignment without being handicapped by insufficient feature matches. Fig.~\ref{fig:overview} gives a preview of our method.

\subsection{Previous work on center insertion}\label{sec:previous}
\vspace{-0.5em}
Center insertion has been studied extensively in spline regression~\cite[Chapter 5]{hastie08}. In particular, center insertion has been proposed for \emph{pixel-based} non-rigid object registration~\cite{cham99,marsland03,bartoli04}. A 2D spline $f:\mathbb{R}^2 \mapsto \mathbb{R}^2$ is a function
\begin{align}\label{equ:spline}
f(\bx) = \bA^T \tilde{\bx} + \sum^{K}_{k=1} \alpha_k \phi(\|\bx - \bc_k \|_2),
\end{align}

where $\bA \in \mathbb{R}^{3 \times 2}$ is an affine warp, $\tilde{\bx} = [\bx^T,1]^T$ is $\bx$ in augmented coordinates, $\{\alpha_k\}$ are scalar coefficients, $\{\bc_k\}$ are 2D positions called centers, and $\phi$ is a radial basis function (RBF). The centers can be arbitrary (e.g., on a grid~\cite{szeliski97} over $\mathbb{R}^2$), and need not coincide with detected features.

The complexity of the warp increases with the number of centers $K$. If the pixels cannot be aligned well due to insufficient warp flexibility, one may consider adding new centers $\bc_\ast$. Each insertion requires deciding where to place $\bc_\ast$, and how to update the parameters $\{\bA,\alpha_1,\dots,\alpha_K, \alpha_\ast\}$. W.r.t.~the latter, in~\cite{bartoli04} the Gauss-Newton algorithm is used to adjust the parameters to further minimize the intensity difference in the overlap area. Note that the spline parameters are not independent, e.g., the coefficients in Thin Plate Splines (TPS) must satisfy the side condition $\sum_{k} \alpha_k = 0$. Thus, the updates get costlier as more centers are inserted.

Note that if $\bx$ is sufficiently far away from all $\{ \bc_k \}$, the side condition and the monotonically decreasing RBF ensures that $f(\bx)$ reduces to the affinity $\bA$. This implies that splines are unsuitable for image stitching, since ideally the warp should revert to a homography in the extrapolation areas~\cite{zaragoza13}. While there exist splines with a projective baseline~\cite{bartoli10}, the fact remains that parameter updating can be relatively non-trivial. We show how the equivalent optimization on MLS regression is much simpler and more efficient.

\section{Correspondence Search}\label{sec:search}
\vspace{-0.5em}
Our goal is to find a warping function $f(\bx)$ that maps pixels from the source image $I$ to the target image $I^\prime$. A set of point-wise matches $\cX = \{\bx_i,\bx^\prime_i \}^{N}_{i=1}$ are first established across $I$ and $I^\prime$, where $\bx_i = [p_i,q_i]^T$ and $\bx^\prime_i = [p^\prime_i,q^\prime_i]^T$. The matches provide a sample of the true underlying warp, and we wish to estimate $f(\bx)$ from $\cX$. In regions where $\cX$ undersamples the true warp (e.g., insufficient point matches), the accuracy of $f(\bx)$ in approximating the true warp is limited. We wish to construct a method to generate new correspondences $\{\bx_\ast,\bx^\prime_\ast\}$ to improve $f(\bx)$, given that $f(\bx)$ is modeled as an APAP warp~\cite{zaragoza13}.

This section describes a novel algorithm to optimize $\bx_\ast^\prime$ for a newly inserted $\bx_\ast$. Sec.~\ref{sec:insert} presents a data-driven algorithm for choosing $\bx_\ast$, in the context of panoramic stitching.

\subsection{APAP warp}

\vspace{-0.5em}

An APAP warp is defined by
\begin{align}\label{equ:apap}
f(\bx) = \left[ \frac{\bh_1(\bx)^T \tilde{\bx}}{\bh_3(\bx)^T \tilde{\bx}}~~,~~\frac{\bh_2(\bx)^T \tilde{\bx}}{\bh_3(\bx)^T \tilde{\bx}} \right]^T,
\end{align}
which is basically a projective warp (here, defined in inhomogeneous coordinates), but where the homography
\begin{align}\label{equ:inputdep}
\bH(\bx) = [\bh_1(\bx),\bh_2(\bx),\bh_3(\bx)]^T \in \mathbb{R}^{3 \times 3}
\end{align}
is input-dependent, and $\bH(\bx)$ is estimated using MDLT as
\begin{align}\label{equ:mdlt}
\bh(\bx) = \argmin_{\bh} \sum^N_{i=1} w_i(\bx) \| \bm_i \bh  \|^2_2,~~~~\text{s.t.}~~\|\bh\|=1.
\end{align}
Here, $\bh(\bx) \in \mathbb{R}^9$ is the column-wise vectorized form of $\bH(\bx)$, $\bm_i \in \mathbb{R}^{2 \times 9}$ contains monomials from linearizing the homography constraint for the $i$-th datum $\{\bx_i,\bx^\prime_i \}$
\begin{align}\label{equ:monomials}
\bm_i = \left[ \begin{matrix} \mathbf{0}_{1\times 3} & -\tilde{\bx}^T_i & q^\prime_i \tilde{\bx}^T_i \\ \tilde{\bx}^T_i & \mathbf{0}_{1 \times 3} & -p^\prime_i \tilde{\bx}^T_i \end{matrix} \right],
\end{align}
and $w_i(\bx)$ is a \emph{non-stationary} weight
\begin{align}\label{equ:weight}
w_i(\bx) = \exp(-\| \bx - \bx_i \|^2_2/2\sigma^2).
\end{align}
Intuitively, \eqref{equ:apap} is a \emph{moving average} of locally weighted projective warps, where $\sigma$ in~\eqref{equ:weight} controls the warp smoothness.

Equation~\eqref{equ:mdlt} defines a weighted algebraic least squares problem, which can be rewritten in the matrix form
\begin{align}\label{equ:mdlt2}
\bh(\bx) = \argmin_\bh \left\| \bW(\bx) \bM \bh \right\|^2_2,~~~~\text{s.t.}~~\|\bh\|=1,
\end{align}
where $\bW(\bx)$ is a $2N \times 2N$ diagonal matrix containing
\begin{align}
w_1(\bx),w_1(\bx),w_2(\bx),w_2(\bx),\dots,w_N(\bx),w_N(\bx),
\end{align}
and $\bM$ is a $2N \times 9$ matrix obtained by vertically stacking the monomial matrices $\bm_1,\bm_2,\dots,\bm_N$. The solution to~\eqref{equ:mdlt2} is the least significant eigenvector of $[\bW(\bx)\bM]^T\bW(\bx)\bM$.

\subsection{Objective function and minimization}

\vspace{-0.5em}
In regions with sparse correspondences, $\bx$ is equally far (relative to $\sigma$) from all $\{\bx_i \}^{N}_{i=1}$, and $f(\bx)$ reduces to a ``rigid" projective warp, thus losing its spatially varying ability. Let $\bx_\ast$ be a newly inserted point in $I$ to raise the flexibility of $f(\bx)$ (again, selecting $\bx_\ast$ will be discussed in Sec.~\ref{sec:insert}). In the absence of geometric information, we need to rely on pixel intensity values to find a matching point $\bx^\prime_\ast$ in $I^\prime$. To this end, we define the intensity matching cost
\begin{align}\label{equ:obj}
E(\bx^\prime_\ast) = \sum_{\bx \in \mathbb{D}} \left[ I^\prime(f(\bx|\bx^\prime_\ast)) - I(\bx) \right]^2,
\end{align}
where $I(\bx)$ is the pixel intensity at $\bx$, $\mathbb{D}$ is a region in $I$ (by default, $\mathbb{D}$ is a $31 \times 31$ subwindow), and the warp $f(\bx|\bx_\ast^\prime)$ is now dependent on $\bx^\prime_\ast$. Specifically, the input dependent homography is now obtained as
{\small
\begin{gather}
\nonumber \bh(\bx|\bx^\prime_\ast) = \argmin_{\bh} \sum^N_{i=1} w_i(\bx) \| \bm_i \bh  \|^2_2 + w_\ast(\bx) \| \bm_\ast(\bx^\prime_\ast) \bh \|^2_2,\\
\text{s.t.}~\| \bh \| = 1,\label{equ:mdlt3}
\end{gather}}
\hspace{-0.55em} where $\bm_\ast(\bx^\prime_\ast)$ contains the monomials for $\{\bx_\ast,\bx^\prime_\ast \}$, and
\begin{align}
w_\ast(\bx) = \exp(-\| \bx - \bx_\ast \|^2_2/2\sigma^2).
\end{align}
In matrix form,~\eqref{equ:mdlt3} can be rewritten as
\begin{gather}
\nonumber \bh(\bx|\bx^\prime_\ast) = \argmin_\bh \left\| \bW_\ast(\bx) \bM(\bx^\prime_\ast) \bh \right\|^2_2,\\
\text{s.t.}~~\|\bh\|=1,\label{equ:mdlt4}
\end{gather}
where $\bW_\ast(\bx)$ is $\bW(\bx)$ diagonally extended with two $w_\ast(\bx)$ values, and $\bM(\bx^\prime_\ast)$ is $\bM$ vertically appended with $\bm_\ast(\bx^\prime_\ast)$. Note that only $\bM(\bx^\prime_\ast)$ contains the variable $\bx^\prime_\ast$.

Our aim is to find $\bx^\prime_\ast$ by minimizing~\eqref{equ:obj}. We apply the well-known Lucas-Kanade (LK) technique~\cite{baker04}. A first-order Taylor expansion is applied on $E(\bx^\prime_\ast + \Delta\bx^\prime_\ast)$ to yield
\begin{align*}
\sum_{\bx \in \mathbb{D}} \left[  I^\prime(f(\bx|\bx^\prime_\ast)) + \nabla I^\prime(f(\bx|\bx^\prime_\ast)) \frac{\partial f(\bx|\bx^\prime_\ast)}{\partial \bx^\prime_\ast} \Delta \bx^\prime_\ast - I(\bx) \right],
\end{align*}
where image gradient $\nabla I^\prime$ is computed using finite differencing. Differentiating against $\bx^\prime_\ast$ and equating to $0$ yields
{\small
\begin{align}\label{equ:update}
\Delta \bx^\prime_\ast = \bF^{-1} \sum_{\bx \in \mathbb{D}} \left[  \nabla I^\prime(f(\bx|\bx^\prime_\ast)) \frac{\partial f(\bx|\bx^\prime_\ast)}{\partial \bx^\prime_\ast}  \right]^T\left[ I(\bx) - I^\prime(f(\bx|\bx^\prime_\ast)) \right]
\end{align}}
\hspace{-0.3em} where $\bF$ is the (approximated) Hessian
{\small
\begin{align*}
\bF = \sum_{\bx \in \mathbb{D}} \left[ \nabla I^\prime(f(\bx|\bx^\prime_\ast)) \frac{\partial f(\bx|\bx^\prime_\ast)}{\partial \bx^\prime_\ast} \right]^T \left[ \nabla I^\prime(f(\bx|\bx^\prime_\ast)) \frac{\partial f(\bx|\bx^\prime_\ast)}{\partial \bx^\prime_\ast}\right].
\end{align*}}
\hspace{-0.6em} The current value for $\bx^\prime_\ast$ is then updated by $\Delta \bx^\prime_\ast$, and the steps are repeated until convergence. Refer to~\cite{baker04} for details.

To initialize $\bx^\prime_\ast$, we map $\bx_\ast$ with the $f(\bx)$ prior to correspondence insertion. It is crucial to note that $\bh(\bx|\bx^\prime_\ast)$ changes for different $\bx \in \mathbb{D}$. Essentially a unique homography is estimated for each $\bx \in \mathbb{D}$ given $\bx^\prime_\ast$, thus realizing a spatially varying warp. This differs from the standard LK approach for ``frame global" projective registration~\cite{baker04}.

\vspace{-1.0em}

\paragraph{Brightness constancy assumption}

The objective function~\eqref{equ:obj} assumes brightness constancy. In our context, this means corresponding pixels across input images have the same colour/brightness. This assumption may not hold in general, especially if the images are taken using cameras with various color auto-correction routines. To ensure the applicability of our method, we can apply color normalization techniques on the input images prior to stitching~\cite{xu10}.

\subsection{Jacobian of APAP warp}
\vspace{-0.5em}

Evaluating $f(\bx|\bx^\prime_\ast)$ and its Jacobian requires solving the weighted algebraic least squares problem~\eqref{equ:mdlt4} at each iteration - again, this differs from the common types of parametric motions used in LK~\cite{baker04}. Specifically, the solution $\bh(\bx|\bx^\prime_\ast)$ to~\eqref{equ:mdlt4} is the least significant eigenvector of
\begin{align}\label{equ:matrix}
\bS(\bx|\bx^\prime_\ast) := [\bW_\ast(\bx)\bM(\bx^\prime_\ast)]^T[\bW_\ast(\bx)\bM(\bx^\prime_\ast)],
\end{align}
where $\bS(\bx|\bx^\prime_\ast)$ varies with $\bx^\prime_\ast$. The eigenvector satisfies
\begin{gather}
\left[\bS(\bx|\bx^\prime_\ast) - \lambda(\bx|\bx^\prime_\ast)\right]\bh(\bx|\bx^\prime_\ast) = 0,\\
\| \bh(\bx|\bx^\prime_\ast) \| = 1,
\end{gather}
where $\lambda(\bx|\bx^\prime_\ast)$ is the eigenvalue. Via the chain rule,
\begin{align}\label{equ:jacobian}
\frac{\partial f(\bx|\bx^\prime_\ast)}{\partial \bx^\prime_\ast} = \frac{\partial f(\bx|\bx^\prime_\ast)}{\partial \bh(\bx|\bx^\prime_\ast)} \frac{\partial \bh(\bx|\bx^\prime_\ast)}{\partial \bx^\prime_\ast}.
\end{align}
The first term can be obtained by differentiating~\eqref{equ:apap} - for brevity, we do not describe this simple process here.

The second term requires differentiating the eigenvector. Based on known results~\cite{magnus85}, the following expression
{\small
\begin{align}
\frac{\partial \bh(\bx|\bx^\prime_\ast)}{\partial \bx^\prime_\ast} &= \left[ \lambda(\bx|\bx^\prime_\ast) \mathbf{I} - \bS(\bx|\bx^\prime_\ast) \right]^{\dagger} \frac{\partial \bS(\bx|\bx^\prime_\ast)}{\partial \bx^\prime_\ast} \bh(\bx|\bx^\prime_\ast)
\end{align}}
\hspace{-0.5em} can be derived, where $\mathbf{I}$ is the identity matrix. The derivative of $\bS(\bx|\bx^\prime_\ast)$ can in turn be obtained based on~\eqref{equ:matrix} - again, for brevity, we do not describe this simple process here. Note that only the last-two rows of $\bM(\bx^\prime_\ast)$ depend on $\bx^\prime_\ast$.

Our correspondence search procedure is summarized in Algorithm~\ref{alg:lk}. Note that in Step~\ref{step:eig}, the eigenvector $\bh(\bx|\bx^\prime_\ast)$ for each $\bx \in \mathbb{D}$ needs to be calculated. Using modern linear algebra packages, this does not represent significant computational load, even for large $\bS(\bx|\bx^\prime_\ast)$, e.g., $1000 \times 1000$. Moreover, an incremental decomposition scheme~\cite{zaragoza13} can be used to further reduce computational cost.

\begin{algorithm}[t]\centering
\begin{algorithmic}[1]
\REQUIRE Images $I$ and $I^\prime$, feature matches $\{\bx_i,\bx^\prime_i \}^{N}_{i=1}$, novel point $\bx_\ast$.
\STATE Initialize $\bx^\prime_\ast$ by warping $\bx_\ast$ using~\eqref{equ:apap}.
\REPEAT
\FOR{each $\bx \in \mathbb{D}$}
\STATE Solve~\eqref{equ:mdlt4} to obtain $\bh(\bx|\bx^\prime_\ast)$ and $\lambda(\bx|\bx^\prime_\ast)$.\label{step:eig}
\STATE Calculate transformation $f(\bx|\bx^\prime_\ast)$.
\STATE Calculate warp Jacobian~\eqref{equ:jacobian} for $\bx$.
\ENDFOR
\STATE Calculate $\Delta \bx^\prime_\ast$~\eqref{equ:update} and update $\bx^\prime_\ast \leftarrow \bx^\prime_\ast + \Delta \bx^\prime_\ast$.
\UNTIL{$\bx^\prime_\ast$ converges.}
\end{algorithmic}
\caption{Correspondence search for APAP warp.}
\label{alg:lk}
\end{algorithm}

\vspace{-1em}

\paragraph{Comparison against center insertion for splines}

At this juncture, it is instructive to compare our correspondence insertion algorithm with spline-based center insertion techniques (Sec.~\ref{sec:previous}). Our warp update algorithm involves nothing more than searching for a 2D point $\bx^\prime_\ast$. This is a direct consequence of using ``point set surfaces"~\cite{alexa03} to define the warp. Contrast this to spline-based center insertion schemes, where all the warp parameters $\{ \bA, \alpha_1, \dots, \alpha_K, \alpha_\ast \}$ need to be adjusted in each update.


\section{Data-Driven Warp Adaptation}\label{sec:insert}
\begin{figure}
\centering
\subfigure[]{\includegraphics[width = 0.43\linewidth,height = 0.08\textheight]{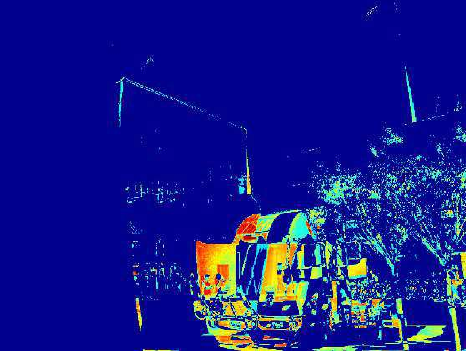}\label{fig:K}}
\hspace{0.3cm}
\subfigure[]{\includegraphics[width = 0.43\linewidth,height = 0.08\textheight]{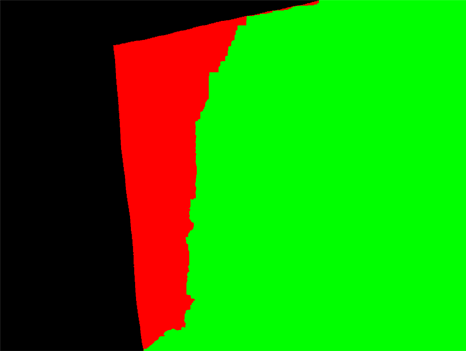}\label{fig:seamcut}}\\
\vspace{-0.5em}
\subfigure[]{\includegraphics[width = 0.43\linewidth,height = 0.08\textheight]{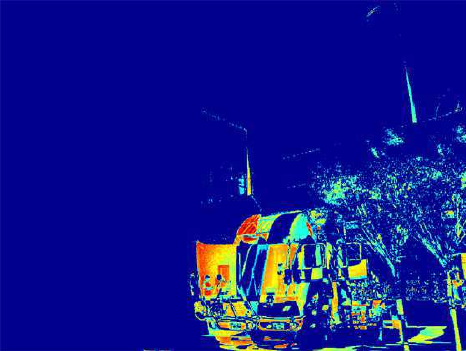}\label{fig:K-sc}}
\hspace{0.3cm}
\subfigure[]{\includegraphics[width = 0.43\linewidth,height = 0.08\textheight]{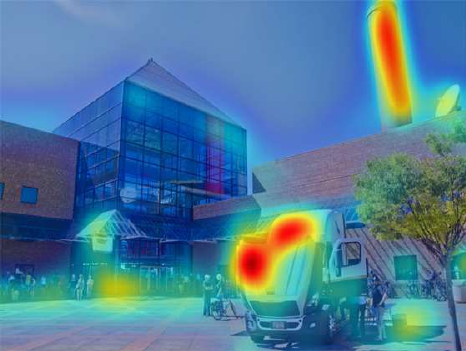}\label{fig:saliency}}\\
\vspace{-0.5em}
\subfigure[]{\includegraphics[width = 0.43\linewidth,height = 0.08\textheight]{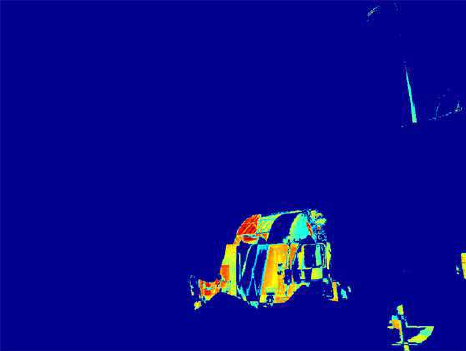}\label{fig:K-salient}}
\hspace{0.3cm}
\subfigure[]{\includegraphics[width = 0.43\linewidth,height = 0.08\textheight]{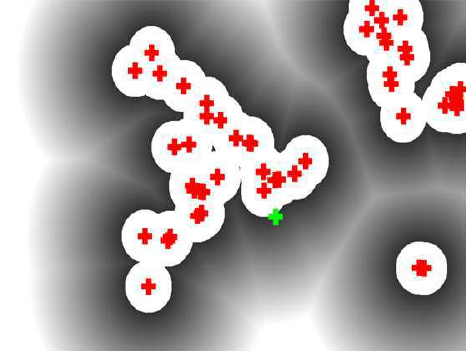}\label{fig:dt}}
\caption{Data-driven warp adaptation. (a) Absolute difference map $R$ with values $< \epsilon$ zeroed; (b) Optimized seam for the current $f(\bx)$; (c) Values in $R$ corresponding to pixels selected from $I^\prime$ are zeroed; (d) Visual saliency map of $I^\prime$; (e) Values in $R$ corresponding to pixels with saliency $< \eta$ zeroed; (f) Distance transform $D$ on $\{ f(\bx_i) \}^{N}_{i=1}$ with values $< \rho$ set to $\infty$ (darker areas here mean lower $D$ values). Green cross indicates the $\bx^\prime_{\min}$ in this iteration.}
\label{fig:centrechoose}
\end{figure}

\vspace{-0.5em}
The previous section presented an algorithm that optimizes $\bx^\prime_\ast$ given $\bx_\ast$. The remaining problem now is how to choose $\bx_\ast$ to improve an APAP warp for image stitching. Our technique is encapsulated in a data-driven warp adaption scheme, which iteratively inserts new correspondences until sufficient ``coverage" of the overlap area is achieved. Algorithm~\ref{alg:centrechoose} summarises the method while Fig.~\ref{fig:centrechoose} illustrates the core steps. Details are in the following.

\begin{algorithm}[t]\centering
\begin{algorithmic}[1]
\REQUIRE Input images $I$ and $I^\prime$, initial correspondence set $\cX=\{\bx_i,\bx^\prime_i \}^{N}_{i=1}$, error threshold $\epsilon$, saliency threshold $\eta$, distance threshold $\rho$, and acceptance threshold $\omega$.
\STATE $\mathcal{L} \leftarrow \{ \bx^\prime_i \}^{N}_{i=1}$.
\STATE Compute visual saliency map on $I^\prime$; see Fig.~\ref{fig:saliency}.
\LOOP
\STATE Estimate APAP warp $f(\bx)$ from $\cX$.
\STATE Warp $I$ to align with $I^\prime$ using $f(\bx)$.
\STATE $R \leftarrow$ absolute intensity diff.~map in overlap area.
\STATE Set values in $R$ which are $< \epsilon$ to $0$; see Fig.~\ref{fig:K}.
\STATE Optimize seam~\cite{agarwala04} for pixel selection in overlap area; see Fig.~\ref{fig:seamcut}.
\STATE Set values in $R$ corresponding to pixels selected from $I^\prime$ according to the seam to $0$; see Fig.~\ref{fig:K-sc}.
\STATE Set values in $R$ corresponding to pixels of $I^\prime$ with saliency $< \eta$ to $0$; see Fig.~\ref{fig:K-salient}.
\STATE $D \leftarrow$ distance transform on $\mathcal{L}$ in the overlap area.
\STATE Set values in $D$ that are $< \rho$ to $\infty$; see Fig.~\ref{fig:dt}.
\STATE If $D ./ R$ is all $\infty$, then break.
\STATE $\bx^\prime_{\min} \leftarrow$ location in $D ./ R$ with minimum value.
\STATE $\bx_\ast \leftarrow f^{-1}(\bx^\prime_{\min})$.
\STATE $\bx_\ast^\prime \leftarrow$ optimized correspondence from Algorithm~\ref{alg:lk}.
\IF{$E(\bx^\prime_\ast) < \omega$}
\STATE $\cX \leftarrow \cX \cup \{\bx_\ast,\bx_\ast^\prime\}$.
\ENDIF
\STATE $\mathcal{L} \leftarrow \mathcal{L} \cup \bx^\prime_{\min}$.
\ENDLOOP
\end{algorithmic}
\caption{Data-driven warp adaptation.}
\label{alg:centrechoose}
\end{algorithm}

Given the current correspondence set $\cX=\{\bx_i,\bx^\prime_i \}^{N}_{i=1}$, an APAP warp $f(\bx)$~\eqref{equ:apap} is first estimated and used to warp the source image $I$ to align with the target image $I^\prime$. Naturally we should strive to add correspondences in regions with high alignment errors. This is provided by the \emph{absolute intensity difference} map $R$. Since we warp $I$ to align with $I^\prime$, it is natural to put $R$ in the same frame as $I^\prime$. We ignore pixels (by zeroing the corresponding values in $R$) with error less than $\epsilon$ (default $\epsilon$ is $100$). See Fig.~\ref{fig:K}.

Our approach relies on seam cut~\cite{agarwala04} for pixel selection during compositing; see Fig.~\ref{fig:seamcut}. Therefore, since pixels that will have their color copied (more appropriately, retained) from $I^\prime$ are not subjected to misalignment errors, the corresponding values in $R$ are zeroed. See Fig.~\ref{fig:K-sc}.

Misalignments in regions with less structured textures (e.g., sky, trees, white board) are less obvious, thus it is less essential to introduce new correspondences in such locations. To realise this intuition, our scheme computes the visual saliency map of $I^\prime$ using the method of~\cite{gbvs}; see Fig.~\ref{fig:saliency}. Values in $R$ corresponding to pixels with saliency less than $\eta$ (default $\eta$ is $0.5$) are zeroed (recall that $R$ has the same coordinate frame as $I^\prime$); see Fig.~\ref{fig:K-salient}.

At this stage, we have now produced an error map to guide the insertion of new correspondences. Additional constraints are given by the existing correspondence set $\cX$. Specifically, we should insert new correspondences in regions that are not too near to $\cX$, so as to avoid inserting redundant correspondences, and also not too far from $\cX$, so as to ensure that correspondence search can be bootstrapped effectively by the existing $f(\bx)$. These constraints are realised by computing the distance transform $D$ on the current set of features $\mathcal{L}$ in $I^\prime$. Values of $D$ that are less than $\rho$ (default $\rho$ is $15$) are set to $\infty$; see Fig.~\ref{fig:dt}.

Given $D$ and $R$, the position $\bx^\prime_{\min}$ that has the lowest value in $D ./ R$ is sought, where ``$./$" indicates element-wise division. The new point $\bx_\ast$ is then obtained as $f^{-1}(\bx^\prime_{\min})$, and Algorithm~\ref{alg:lk} is invoked to find its correspondence $\bx^\prime_\ast$. To calculate the inverse APAP warp $f^{-1}(\bx^\prime_{\min})$, we find the nearest neighbor of $\bx^\prime_{\min}$ in $\{f(\bx_i)\}^{N}_{i=1}$, then warp $\bx^\prime_{\min}$ to $I$ using the inverse $\bH^{-1}(\bx)$ of the input-dependent homography~\eqref{equ:inputdep} of the nearest neighbor point.

The newly inserted correspondence $\{\bx_\ast,\bx^\prime_\ast \}$ is appended to $\cX$, if $E(\bx^\prime_\ast)$ is less than $\omega$ (default $\omega$ is $1000$). Else, the new correspondence is considered unsatisfactory and discarded. In any case, $\bx^\prime_{\min}$ is appended to $\mathcal{L}$ to prevent it from being selected again in the next iteration. The above warp adaptation process is repeated until the overlap area is sufficiently covered by feature correspondences.

As an indication of runtime, invoking Algorithm~\ref{alg:centrechoose} on the image pair in Fig.~\ref{fig:truck} inserted $81$ new correspondences in $65$ seconds, among which $11$ correspondences were accepted.

\section{Results}

\begin{figure*}
\begin{tabular}{cc}
\parbox{0.1\linewidth}{\centering Optical Flow~\cite{liuce09}} &
\parbox{0.85\linewidth}{\includegraphics[width = 1\linewidth,height = 0.1\textheight]{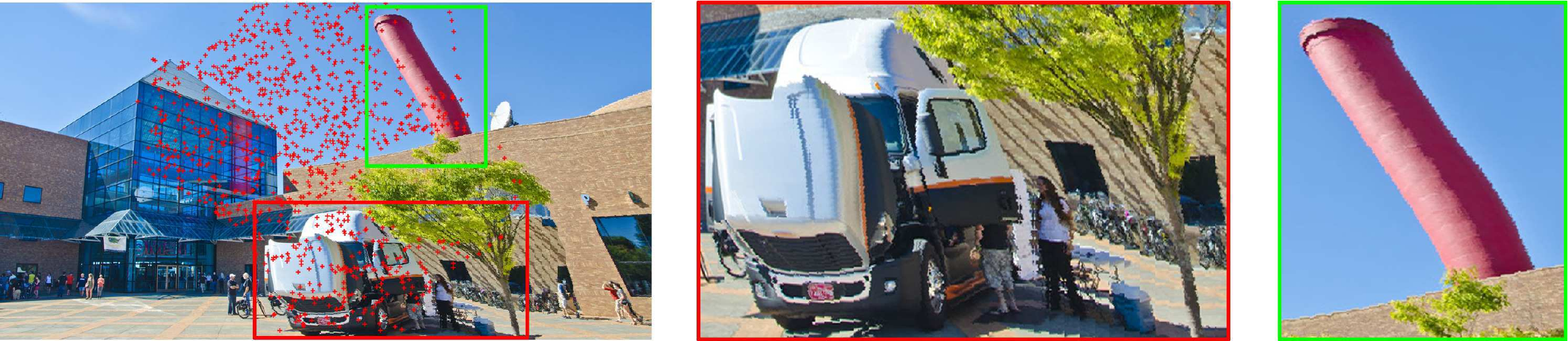}\label{fig:car_of}}\\

\parbox{1.5cm}{\centering Large Displacement Optical Flow~\cite{brox11}} &
\parbox{0.85\linewidth}{\includegraphics[width = 1\linewidth,height = 0.1\textheight]{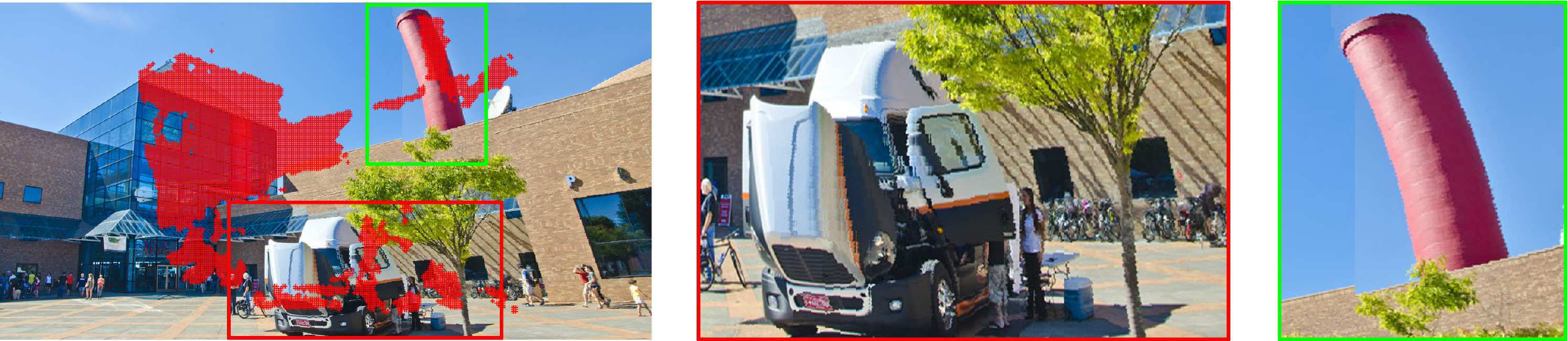}\label{fig:car_tb}}\\

\parbox{1.5cm}{\centering SIFT Flow~\cite{liuce11}} & 
\parbox{0.85\linewidth}{\includegraphics[width = 1\linewidth,height = 0.1\textheight]{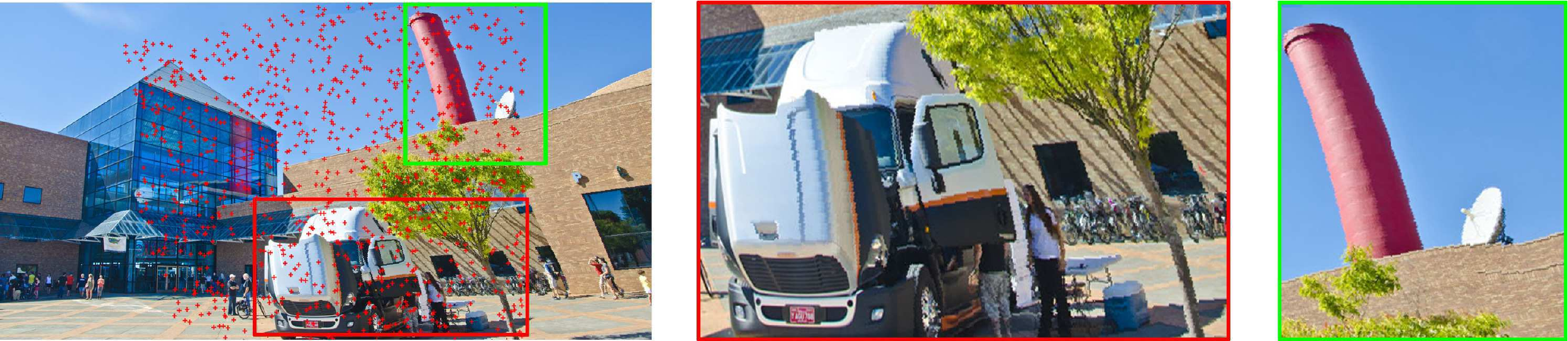}\label{fig:car_sf}}\\
&\\
\end{tabular}
\caption{APAP stitching results using dense correspondences from three flow-based methods. Since optical flow{~\cite{liuce09}} and SIFT flow{~\cite{liuce11}} produce very dense flow fields, to avoid excessive clutter, we display only 1000 randomly sampled correspondences.}
\label{fig:warp_flowbased}
\end{figure*}

\paragraph{Evaluation of flow-based methods}

We used the \emph{truck} image pair from~\cite{zhang14}; see Fig.~\ref{fig:warp_flowbased}. Three state-of-the-art flow-based dense and semi-dense correspondence methods were evaluated~\cite{liuce09,liuce11,brox11}. Before obtaining the dense correspondences, we pre-warped one of the images using a homography estimated from sparse SIFT keypoint matches. This served to simplify the problem for the flow-based methods. Further, RANSAC was invoked with a tight inlier threshold (1 pixel) to ensure high-quality correspondences, before APAP warp~\cite{zaragoza13} (the baseline) was estimated.

Despite the above precautions, the stitching results in Fig.~\ref{fig:warp_flowbased} exhibit significant local distortions. This indicates that many of the correspondences are actually inaccurate. The small error tolerance of RANSAC still allowed sufficient local deviations (e.g., due to repetitive textures) that distorted the warp; see the supplementary material for the actual data used. While such local inaccuracies may not affect motion analysis or segmentation, they are fatal for accurate image stitching using spatially varying warps.

\vspace{-1em}

\paragraph{Comparisons with state-of-the-art stitching methods}

We compared our method (abbreviated as APAP+CI)\footnote{Source code will be made available on our homepages.} against other state-of-the-art approaches, namely the original APAP method~\cite{zaragoza13} and parallax-tolerant image stitching~\cite{zhang14}. We used publicly available images by Zaragoza et al.~and Zhang and Liu, as well as additional images collected by us. Due to space limitations, only a few results can be shown here; see supplementary material for more results.


For APAP warps, we used the code shared by Zaragoza et al. For parallax-tolerant image stitching, we simply reprinted the results (where available) from the project page of Zhang and Liu. For newly collected images, we executed our own implementation of Zhang and Liu's method.

Parameter settings for our method are as follows: $\sigma = 8$ in~\eqref{equ:weight} 
, $\mathbb{D}$ in~\eqref{equ:obj} is a $31 \times 31$ subwindow, $\epsilon = 100$, $\eta = 0.5$, $\rho = 15$, and $\omega = 1000$ in Algorithm~\ref{alg:centrechoose}.

In image pairs with very serious depth parallax, not all pixels have valid correspondences in the other view. Theoretically, the true warping function must ``fold over" or be discontinuous to correctly align the images. Such characteristics are not supported by APAP or the content preserving warp (CPW)~\cite{liu09} used in parallax-tolerant image stitching. Following Zhang and Liu, we thus apply seam cut to composite the images and remove ghosting. 

\begin{figure*}[h]
\centering
\subfigure{\includegraphics[width = 0.9\linewidth,height = 0.143\textheight]{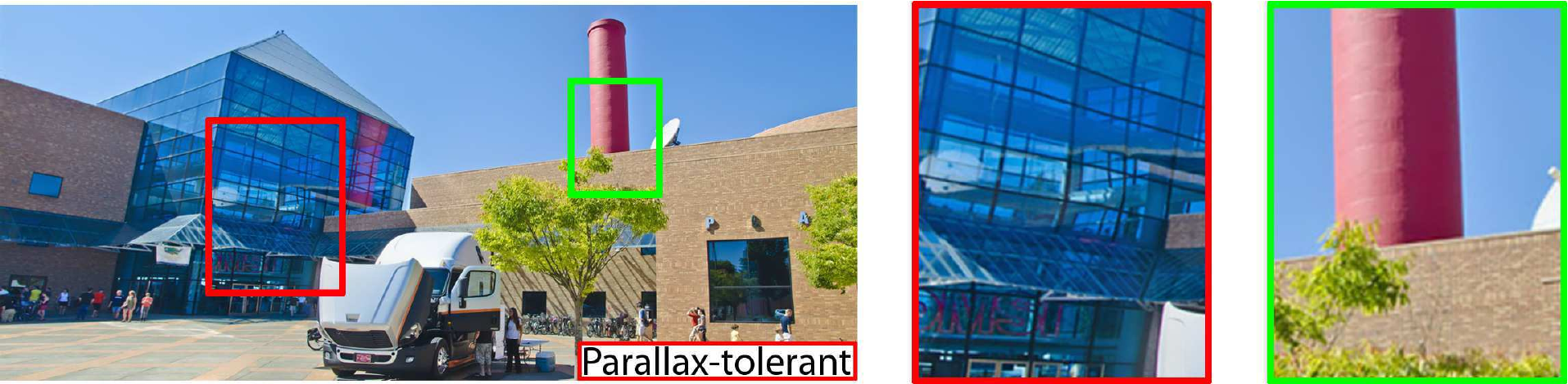}}
\subfigure{\includegraphics[width = 0.9\linewidth,height = 0.143\textheight]{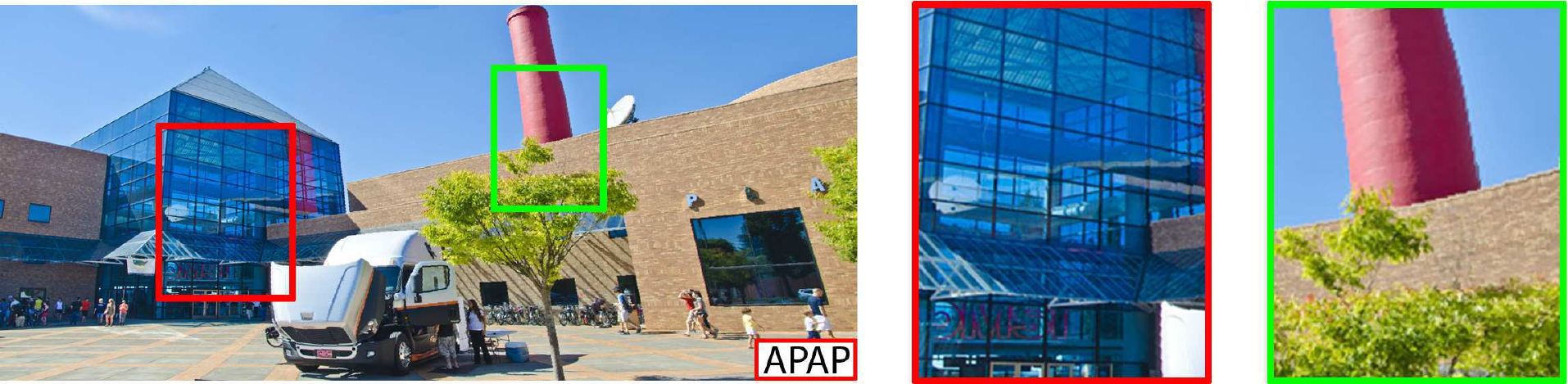}}
\subfigure{\includegraphics[width = 0.9\linewidth,height = 0.143\textheight]{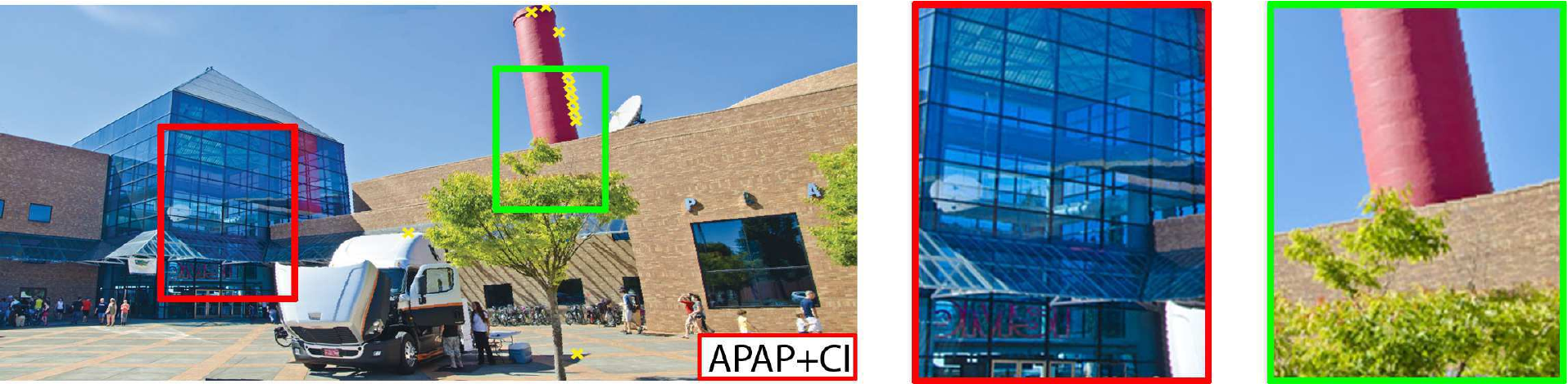}}
\caption{Comparing three methods on \emph{truck} image pair. Inserted correspondences by APAP+CI are shown as yellow points.}
\label{fig:truck}
\end{figure*}

\begin{figure*}[h]
\centering
\subfigure{\includegraphics[width = 0.9\linewidth,height = 0.143\textheight]{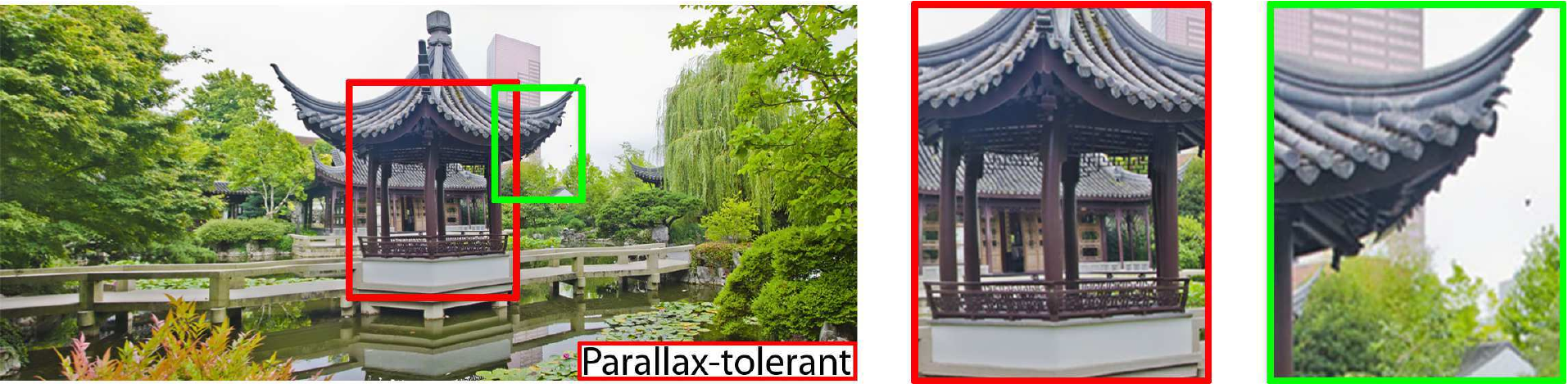}}
\subfigure{\includegraphics[width = 0.9\linewidth,height = 0.143\textheight]{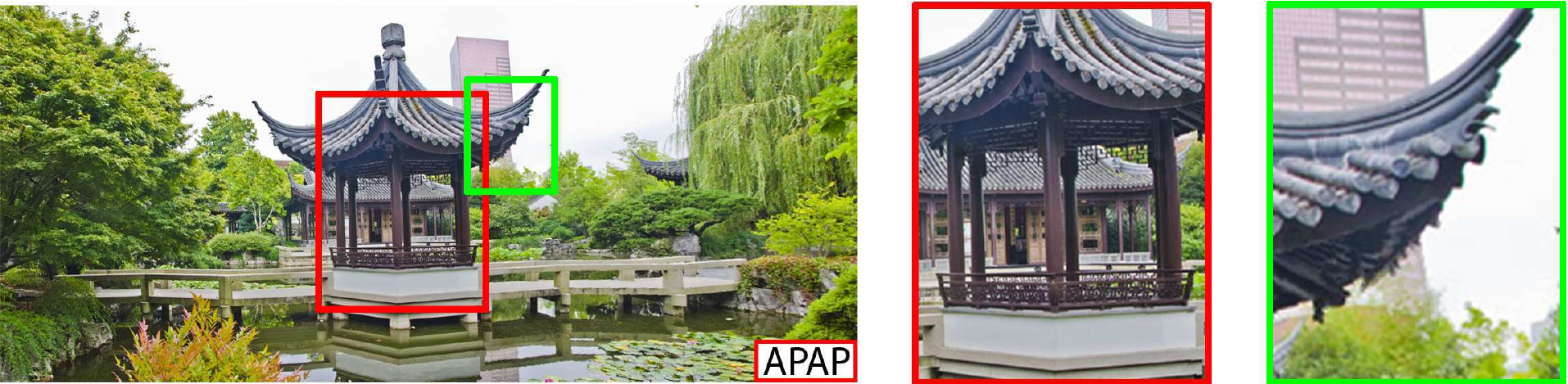}}
\subfigure{\includegraphics[width = 0.9\linewidth,height = 0.143\textheight]{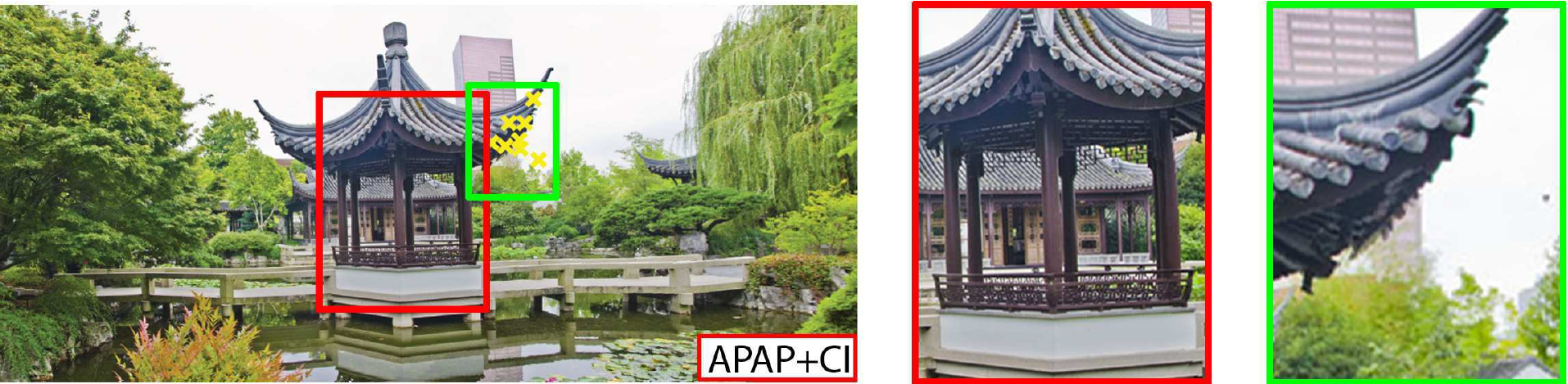}}
\caption{Comparing three methods on \emph{temple} image pair. Inserted correspondences by APAP+CI are shown as yellow points.}
\label{fig:temple}
\end{figure*}

Figs.~\ref{fig:truck} and~\ref{fig:temple} show results on two image pairs used by Zhang and Liu. In Fig.~\ref{fig:truck}, parallax-tolerant image stitching produced significant distortions on the glass building. This was likely due to the concentration of the local homography on the major building to the right, and neglecting the other regions not lying on the same plane (cf.~Fig.~\ref{fig:liufeng}). In contrast, APAP warp was more capable of globally aligning the images; notice that the glass building was not distorted. However, unpleasant distortions exist around the smokestack - due to a lack of feature matches in this region, the warp was ``dragged away" by existing feature matches on the lower building. Our method APAP+CI rectified the distortion by inserting new correspondences in the appropriate positions. In Fig.~\ref{fig:temple}, observe the distortions on the pavilion produced by parallax-tolerant image stitching. Overall, APAP warp accurately aligned the whole image, however, due to the lack of feature correspondences, the tower in the background appeared discontinuous. This was rectified by APAP+CI with the insertion of new correspondences.

\begin{figure*}[h]
\centering
\subfigure{\includegraphics[width = 0.9\linewidth,height = 0.143\textheight]{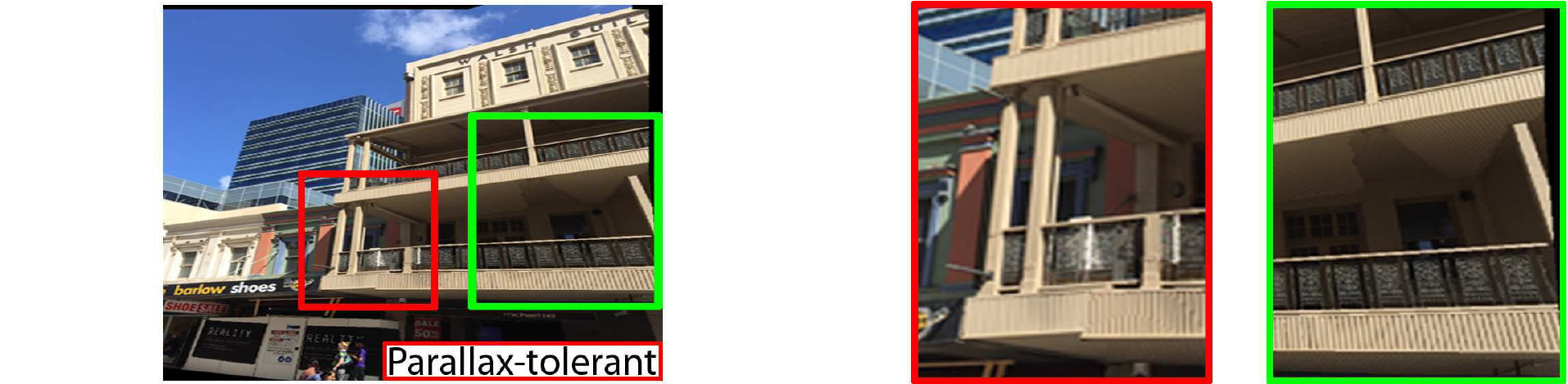}}
\subfigure{\includegraphics[width = 0.9\linewidth,height = 0.143\textheight]{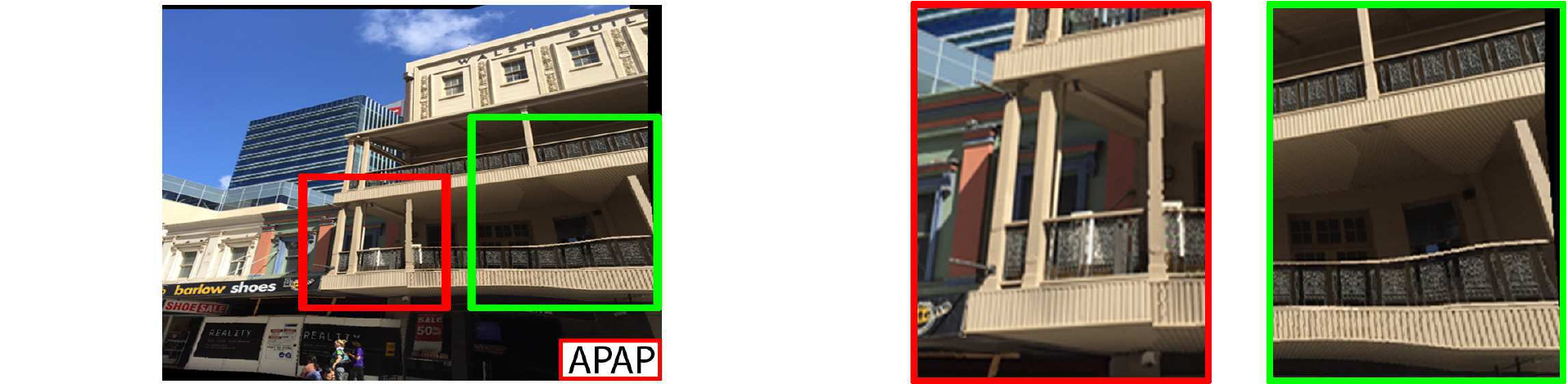}}
\subfigure{\includegraphics[width = 0.9\linewidth,height = 0.143\textheight]{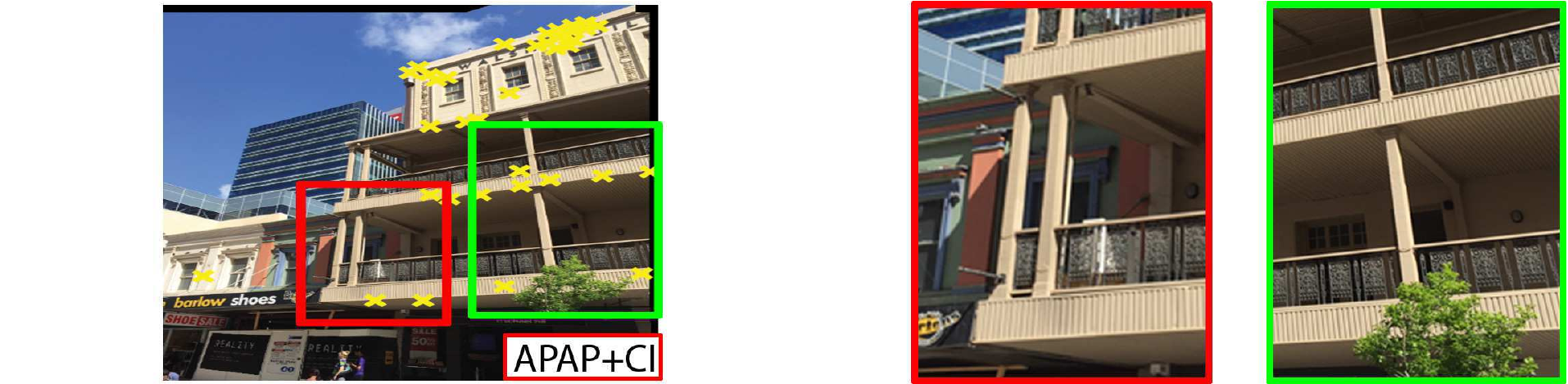}}
\caption{Comparing three methods on \emph{shopfront} image pair. Inserted correspondences by APAP+CI are shown as yellow points.}
\label{fig:rundlemall}
\end{figure*}

\begin{figure*}[h]
\centering
\subfigure{\includegraphics[width = 0.9\linewidth,height = 0.143\textheight]{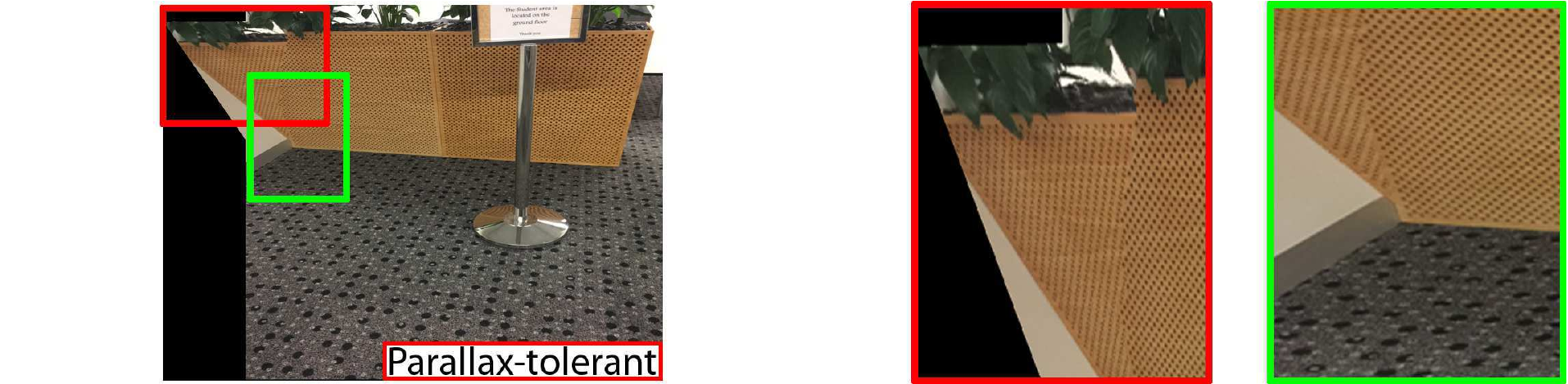}}
\subfigure{\includegraphics[width = 0.9\linewidth,height = 0.143\textheight]{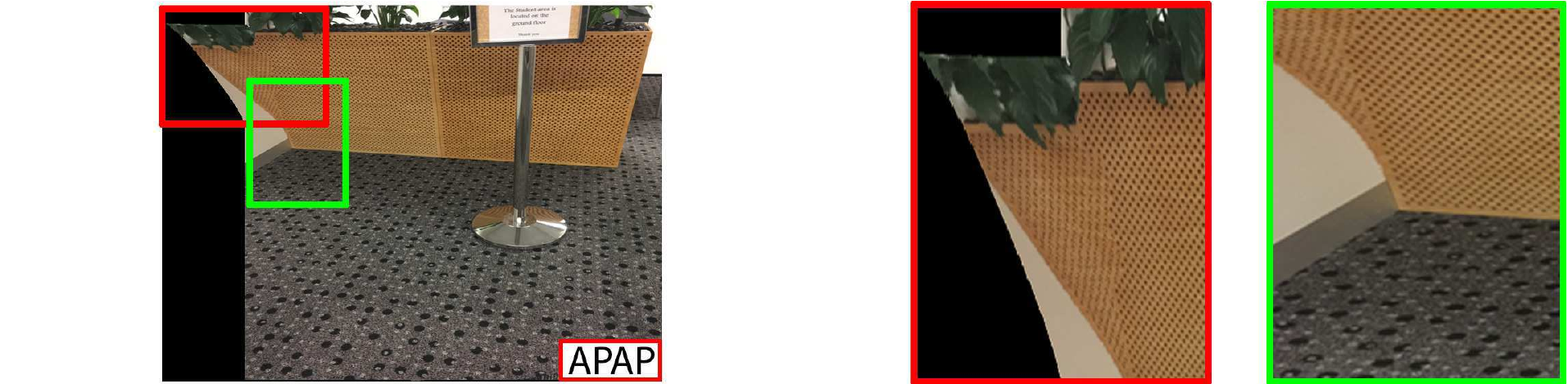}}
\subfigure{\includegraphics[width = 0.9\linewidth,height = 0.143\textheight]{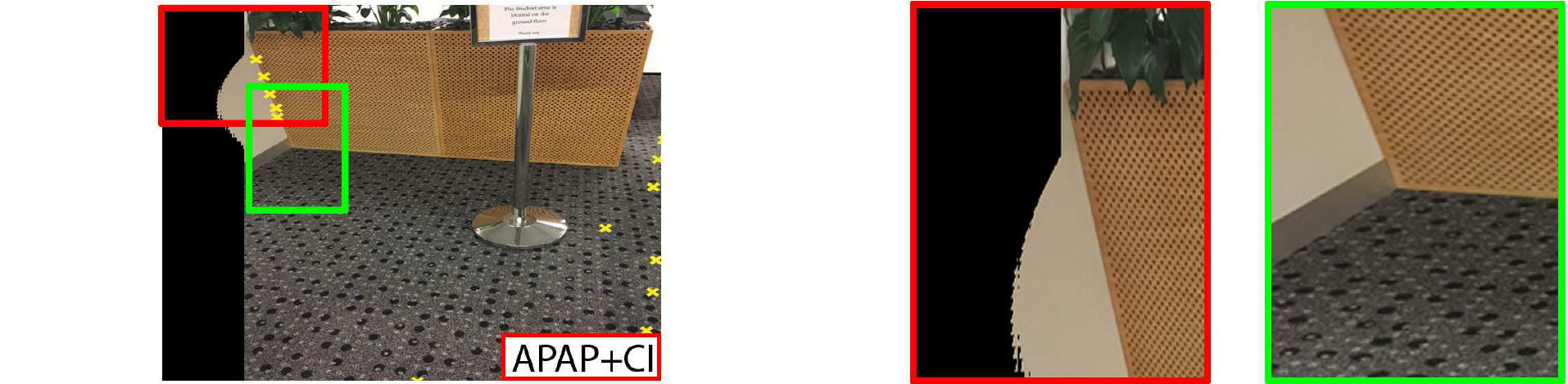}}
\caption{Comparing three methods on \emph{lobby} image pair. Inserted correspondences by APAP+CI are shown as yellow points.}
\label{fig:lvl5}
\end{figure*}

Similar results on two more challenging image pairs are shown in Figs.~\ref{fig:rundlemall} and~\ref{fig:lvl5}; these are newly collected data. Our results show that by inserting new correspondences to adapt the warp, our method rectifies the weakness of APAP warp.

\section{Conclusions}

Wide-baseline image stitching is a challenging problem. Flow-based methods often fail to produce dense \emph{and} accurate correspondences, while spatially varying warps are only flexible up to the sparse set of keypoint matches given. We presented a novel data-driven warp adaption scheme for APAP image stitching. A core step in our algorithm is a correspondence insertion technique. Our method improves upon the original APAP warp, which fails when the overlap region is correspondence-poor. Our results also show that it is crucial to accurately align the images throughout the overlap area, even if sophisticated compositing is used.

\newpage

\clearpage

{\small
\bibliographystyle{ieee}
\bibliography{center_insert}
}

\clearpage
\onecolumn

\begin{center}
  \Large Supplementary Material for\\
Correspondence Insertion for As-Projective-As-Possible Image Stitching
\end{center}
\vskip 5em
\setcounter{section}{0}
\section{Supplementary material overview}

\hspace{0.5cm}This supplementary material provides results on additional image pairs. In particular, Sec.~\ref{sec:flow} shows stitching results using dense correspondences from three flow-based methods~\cite{brox11,liuce09,liuce11}, and Secs.~\ref{sec:parallax} and~\ref{sec:noparallax} compare our novel image stitching method (APAP+CI) against other state-of-the-art methods.

\section{Evaluation of dense correspondences from flow-based methods}\label{sec:flow}

\begin{figure}
\centering
\subfigure[Pre-warp input images using a homography estimated from SIFT keypoint matches.]
{\includegraphics[width = 0.80\linewidth,height = 0.13\textheight]{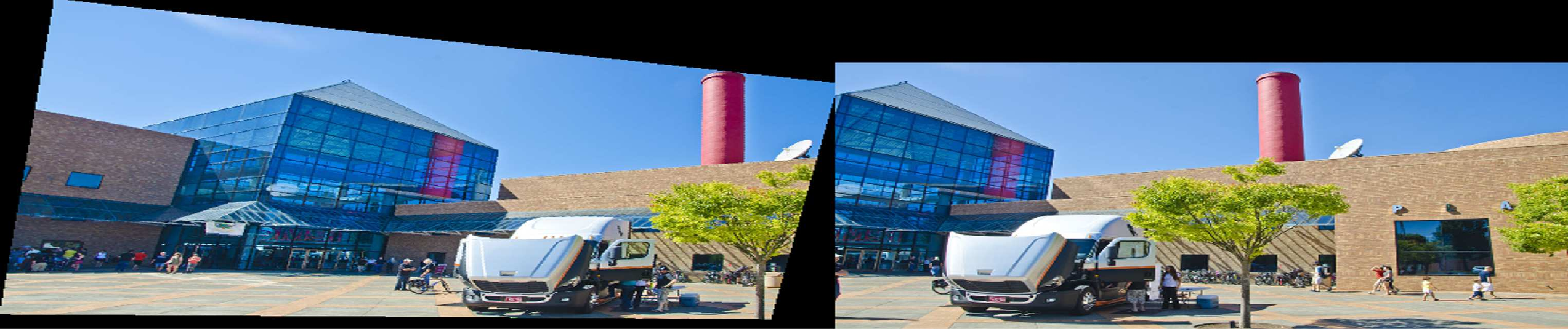}\label{fig1:inputs}}

\subfigure[Image stitching result using APAP warp~\cite{zaragoza13} estimated from a set of semi-dense correspondences (validated by RANSAC and shown as red points in the final stitched image) produced by Large Displacement Optical Flow~\cite{brox11}.]
{\includegraphics[width = 0.80\linewidth,height = 0.13\textheight]{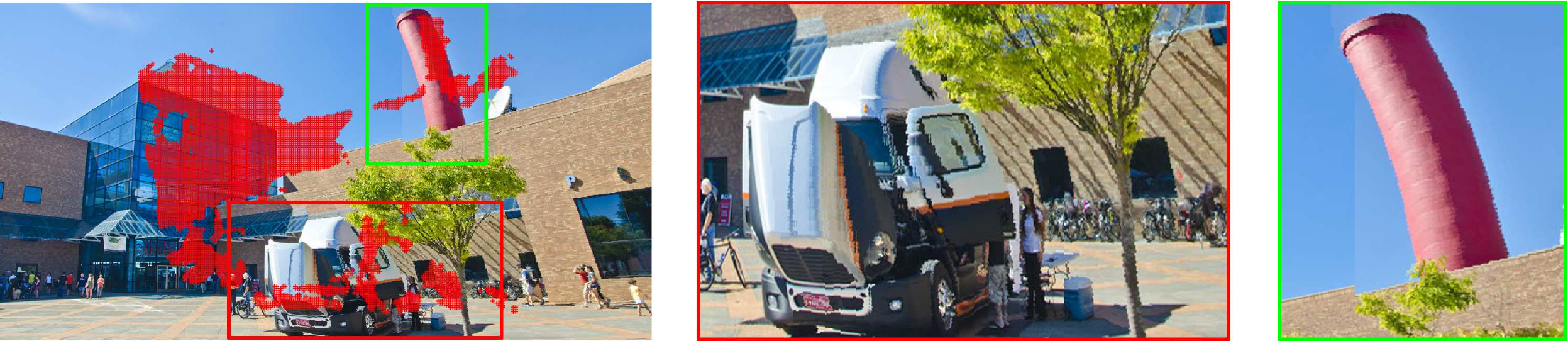}\label{fig1:tb}}

\subfigure[The optic flow implementation of~\cite{liuce09} produced 195868 correspondences (after RANSAC validation).\label{fig1:of_fpts}]
{\includegraphics[width = 0.80\linewidth,height = 0.13\textheight]{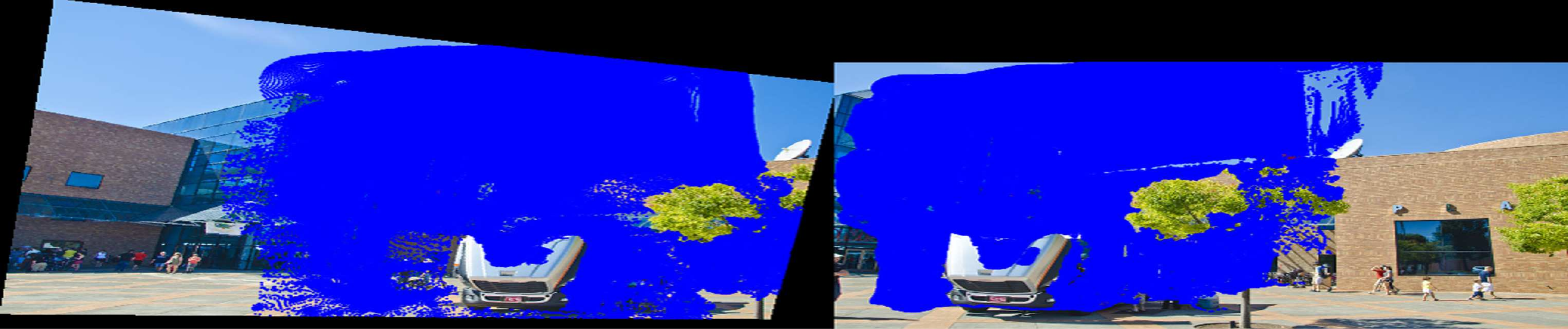}}

\subfigure[Image stitching result using APAP warp~\cite{zaragoza13} estimated from the correspondences in (c).]
{\includegraphics[width = 0.80\linewidth,height = 0.13\textheight]{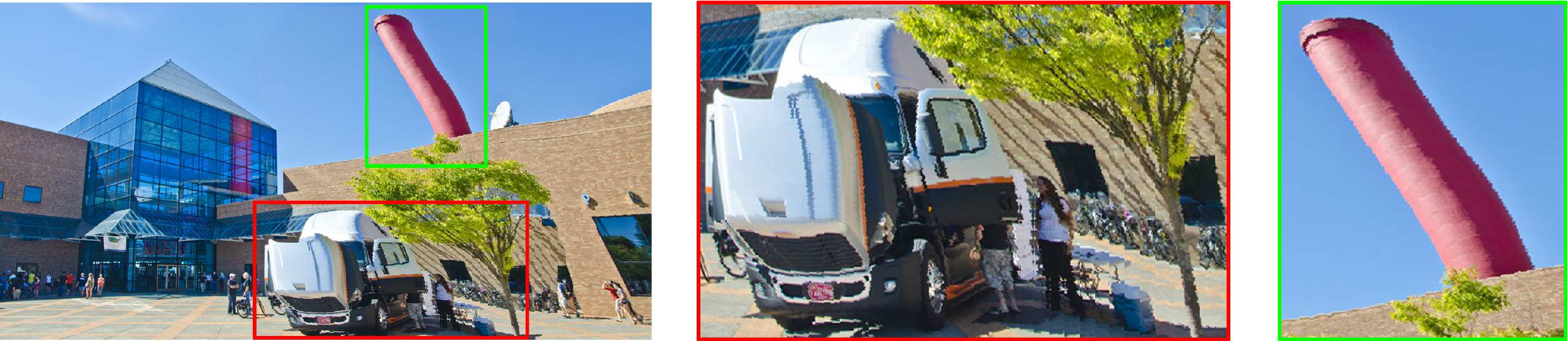}\label{fig1:of}}

\subfigure[SIFT Flow~\cite{liuce11} produced 403308 correspondences (after RANSAC validation).]
{\includegraphics[width = 0.80\linewidth,height = 0.13\textheight]{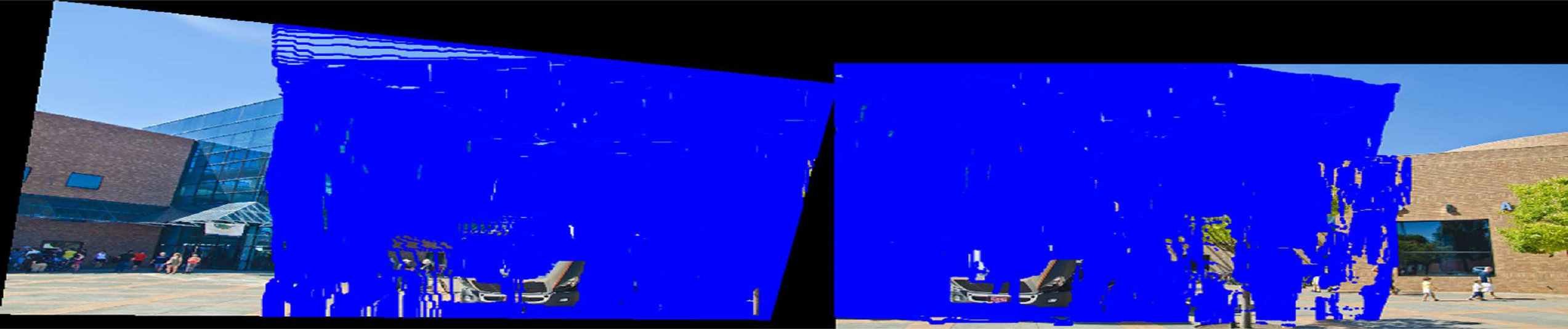}}\label{fig1:sf_fpts}

\subfigure[Image stitching result using APAP warp~\cite{zaragoza13} estimated from the correspondences in (e).]
{\includegraphics[width = 0.80\linewidth,height = 0.13\textheight]{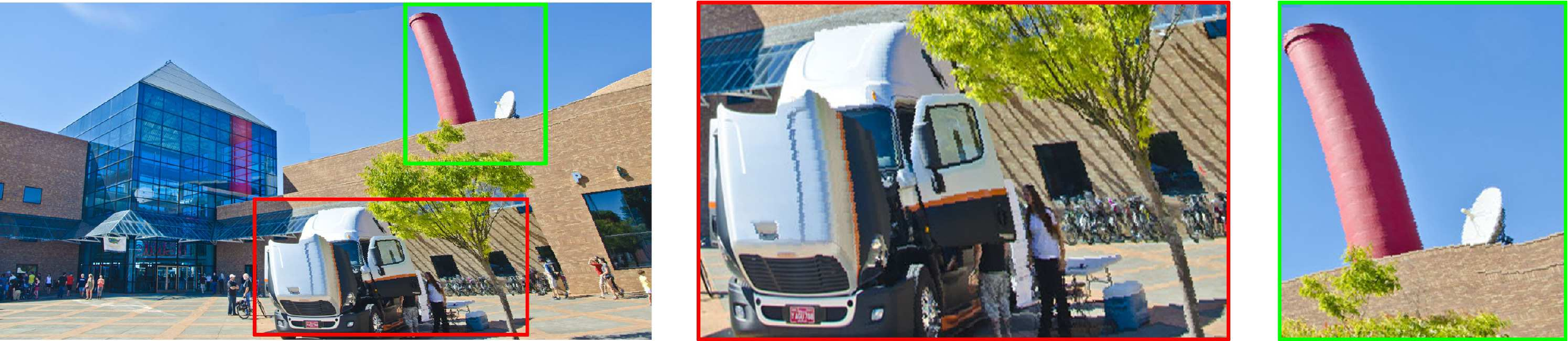}\label{fig1:sf}}
\caption{Dense correspondences and stitching results of three flow-based methods on the~\emph{truck} image pair.}
\label{fig1}
\end{figure}

\begin{figure}
\centering
\subfigure[Pre-warp input images using a homography estimated from SIFT keypoint matches.
\label{fig2:inputs}]
{\includegraphics[width = 0.80\linewidth,height = 0.13\textheight]{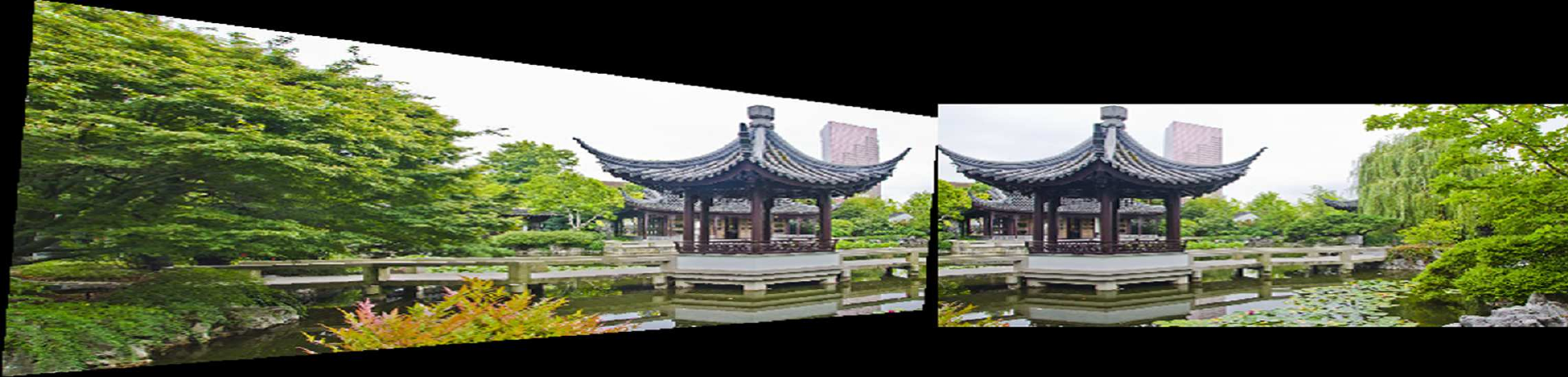}}

\subfigure[Image stitching result using APAP warp~\cite{zaragoza13} estimated from a set of semi-dense correspondences (validated by RANSAC and shown as red points in the final stitched image) produced by Large Displacement Optical Flow~\cite{brox11}.
\label{fig2:tb}]
{\includegraphics[width = 0.80\linewidth,height = 0.13\textheight]{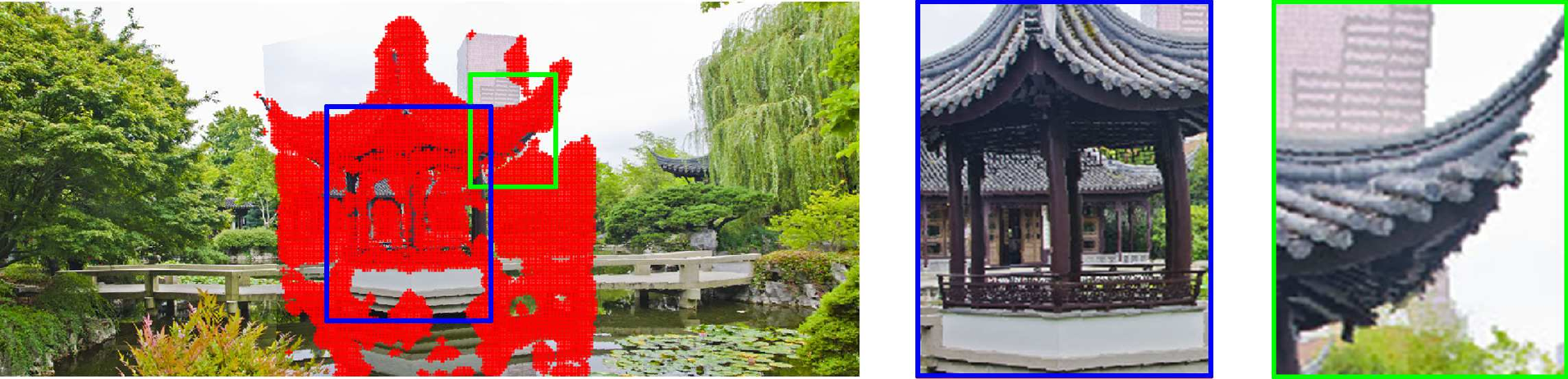}}

\subfigure[The optic flow implementation of~\cite{liuce09} produced 147858 correspondences (after RANSAC validation).
\label{fig2:of_fpts}]
{\includegraphics[width = 0.80\linewidth,height = 0.13\textheight]{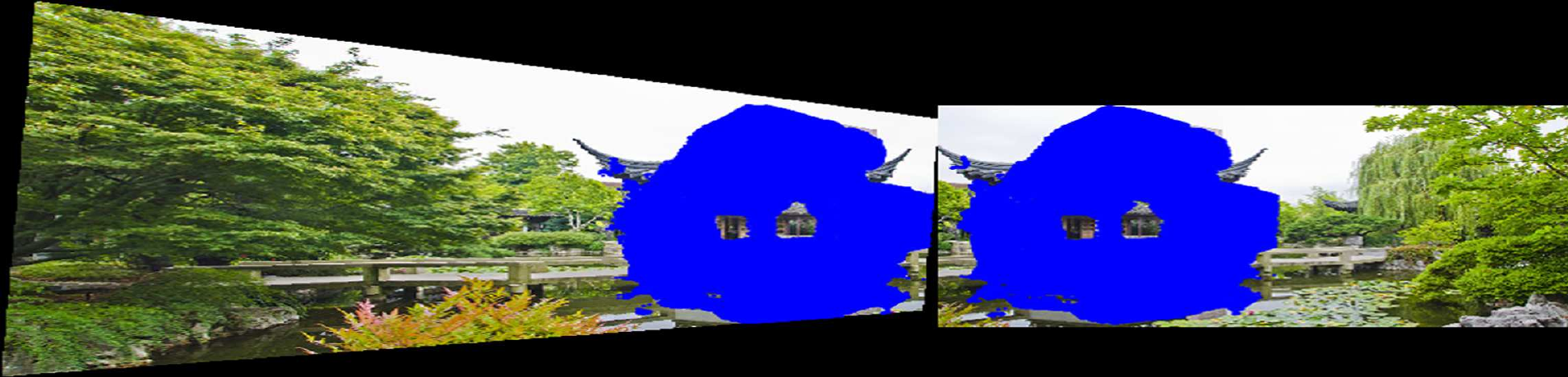}}

\subfigure[Image stitching result using APAP warp~\cite{zaragoza13} estimated from the correspondences in (c).
\label{fig2:of}]
{\includegraphics[width = 0.80\linewidth,height = 0.13\textheight]{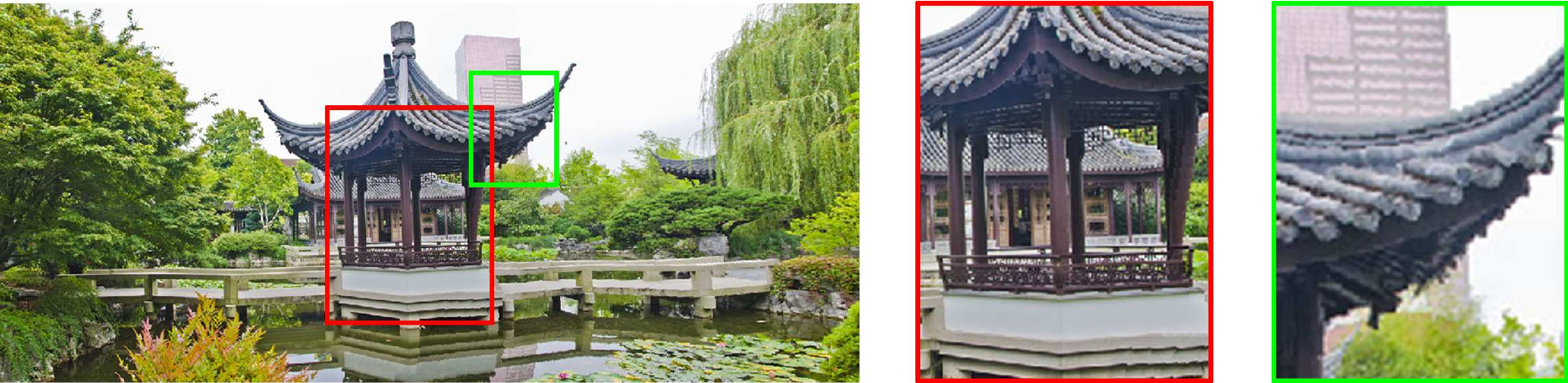}}

\subfigure[SIFT Flow~\cite{liuce11} produced 317753 correspondences (after RANSAC validation).
\label{fig2:sf_fpts}]
{\includegraphics[width = 0.80\linewidth,height = 0.13\textheight]{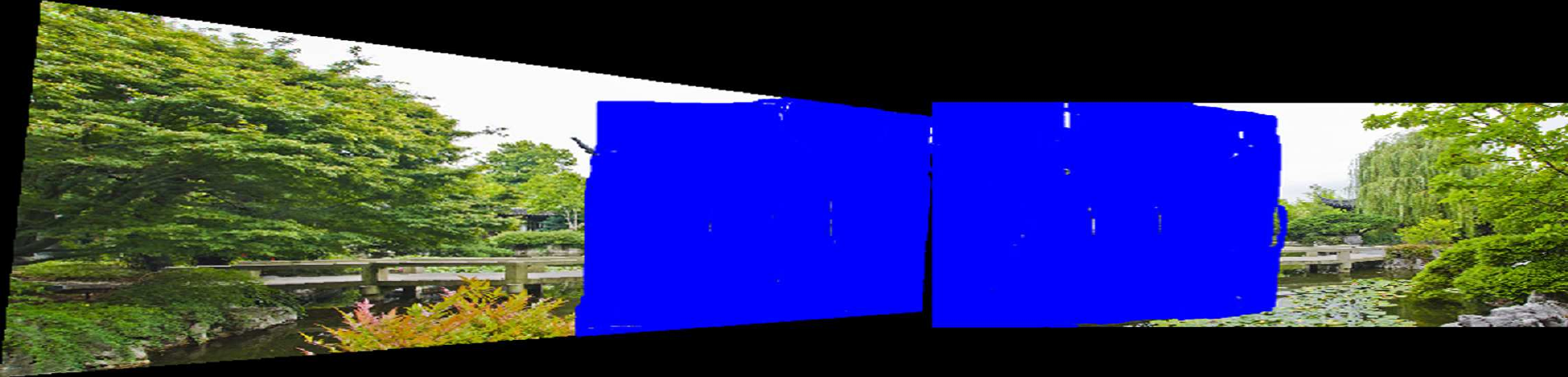}}

\subfigure[Image stitching result using APAP warp~\cite{zaragoza13} estimated from the correspondences in (e).\label{fig2:sf}]
{\includegraphics[width = 0.80\linewidth,height = 0.13\textheight]{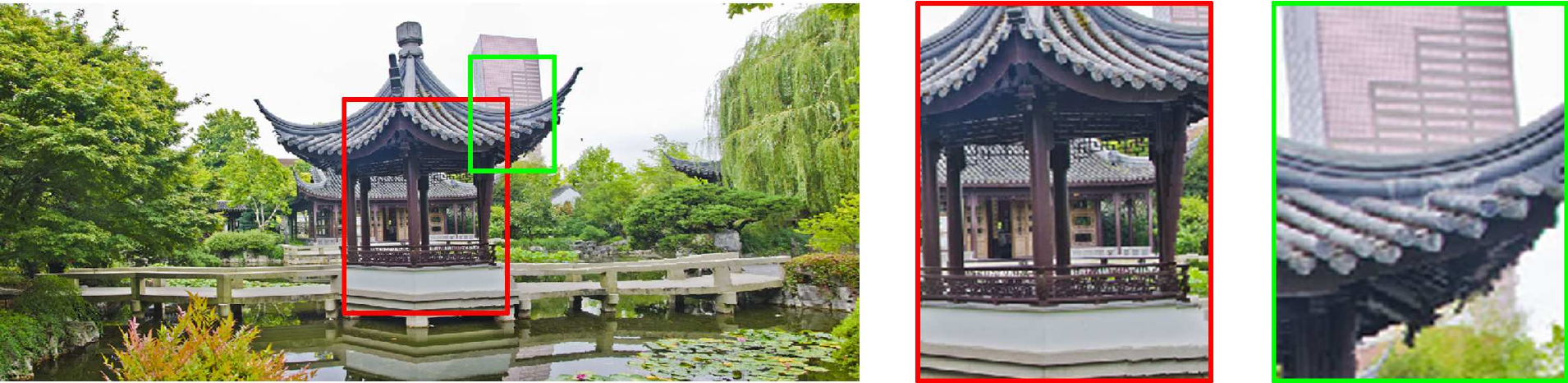}}

\caption{Dense correspondences and stitching results of three flow-based methods on the~\emph{temple} image pair.}
\label{fig2}
\end{figure}

\begin{figure}
\centering
\subfigure[Pre-warp input images using a homography estimated from SIFT keypoint matches.
\label{fig3:inputs}]
{\includegraphics[width = 0.75\linewidth,height = 0.13\textheight]{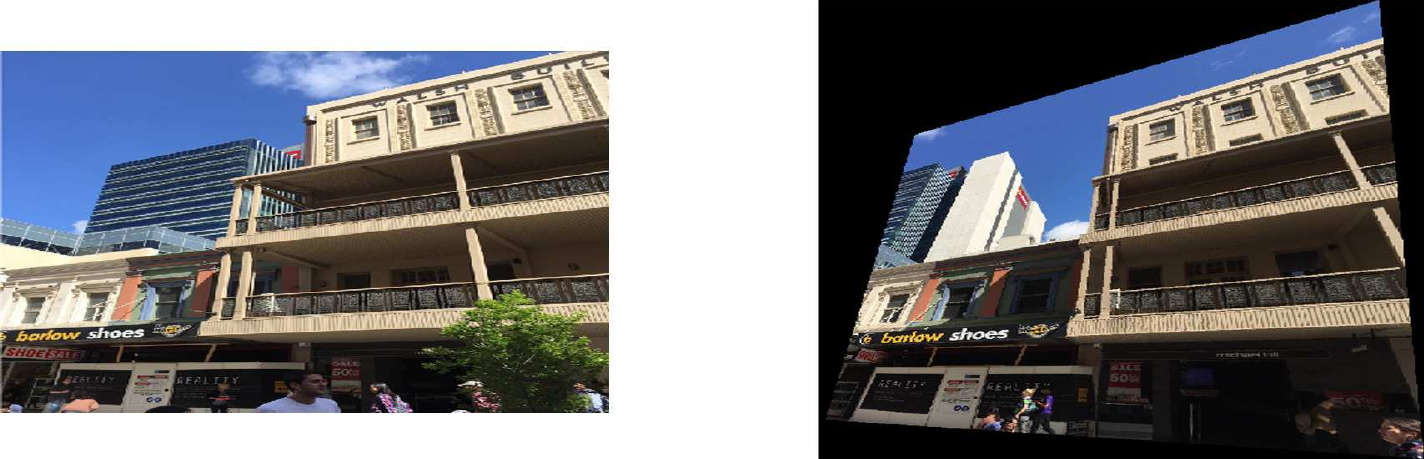}}

\subfigure[Image stitching result using APAP warp~\cite{zaragoza13} estimated from a set of semi-dense correspondences (validated by RANSAC and shown as red points in the final stitched image) produced by Large Displacement Optical Flow~\cite{brox11}.
\label{fig3:tb}]
{\includegraphics[width = 0.75\linewidth,height = 0.13\textheight]{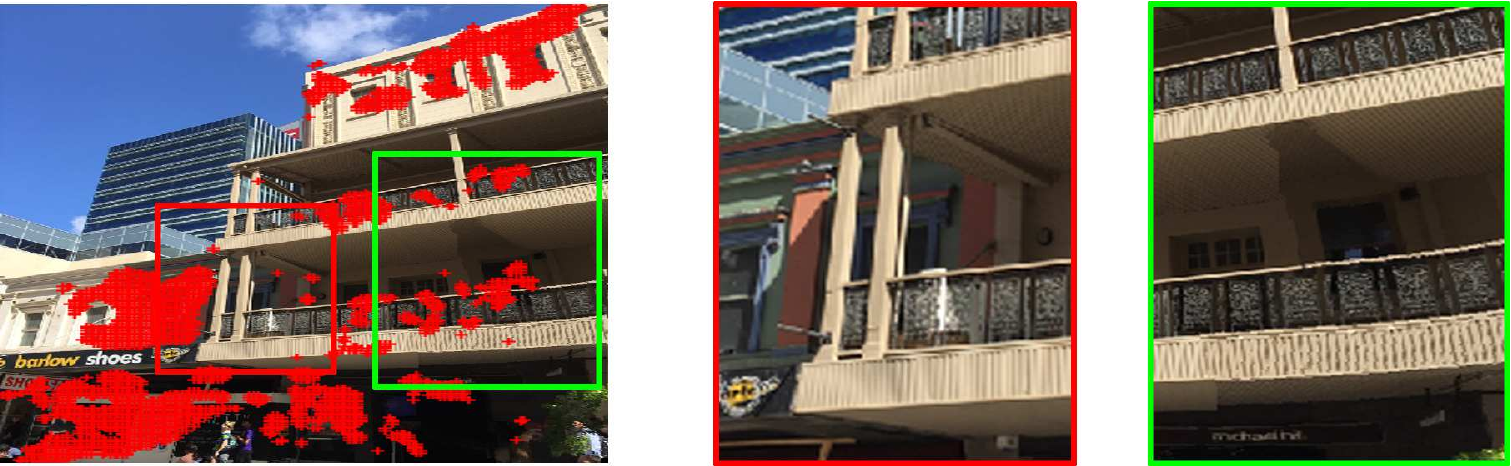}}

\subfigure[The optic flow implementation of~\cite{liuce09} produced 83318 correspondences (after RANSAC validation).
\label{fig3:of_fpts}]
{\includegraphics[width = 0.75\linewidth,height = 0.13\textheight]{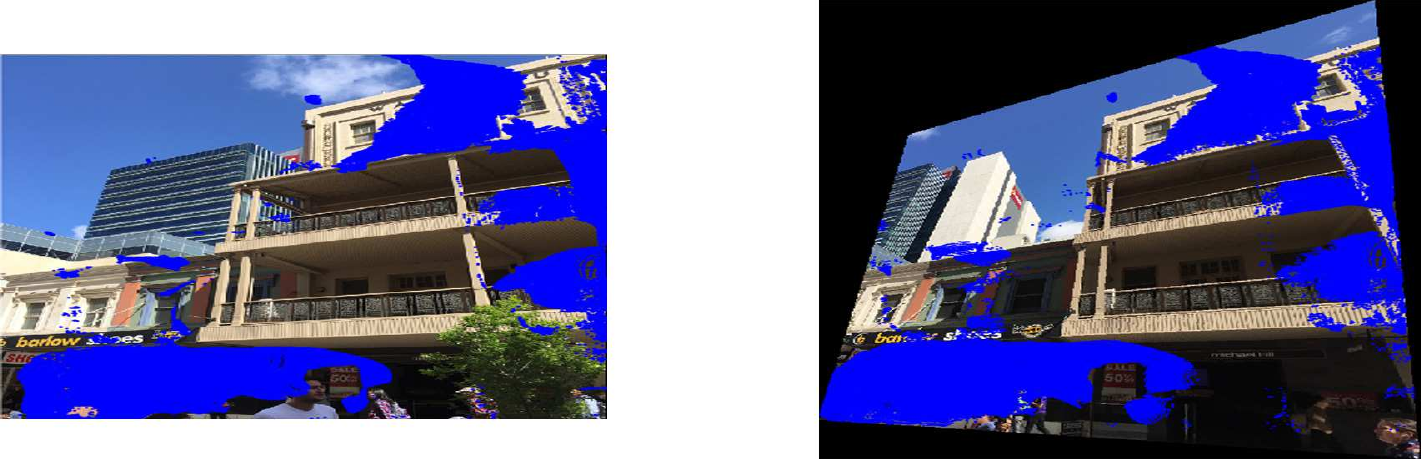}}

\subfigure[Image stitching result using APAP warp~\cite{zaragoza13} estimated from the correspondences in (c).
\label{fig3:of}]
{\includegraphics[width = 0.75\linewidth,height = 0.13\textheight]{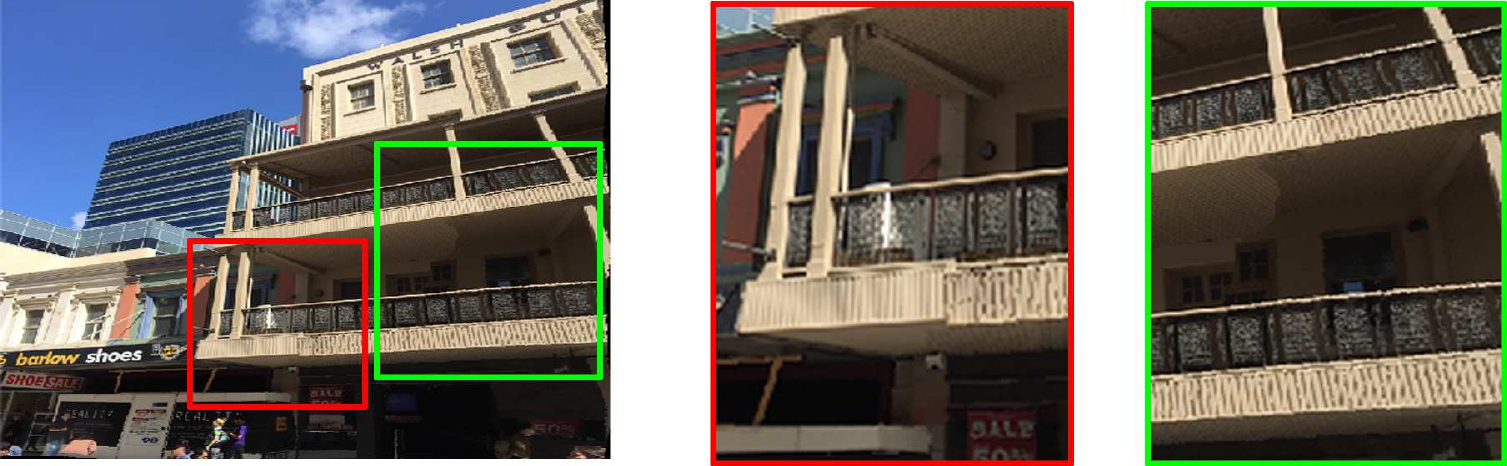}}

\subfigure[SIFT Flow~\cite{liuce11} produced 165789 correspondences (after RANSAC validation).
\label{fig3:sf_fpts}]
{\includegraphics[width = 0.75\linewidth,height = 0.13\textheight]{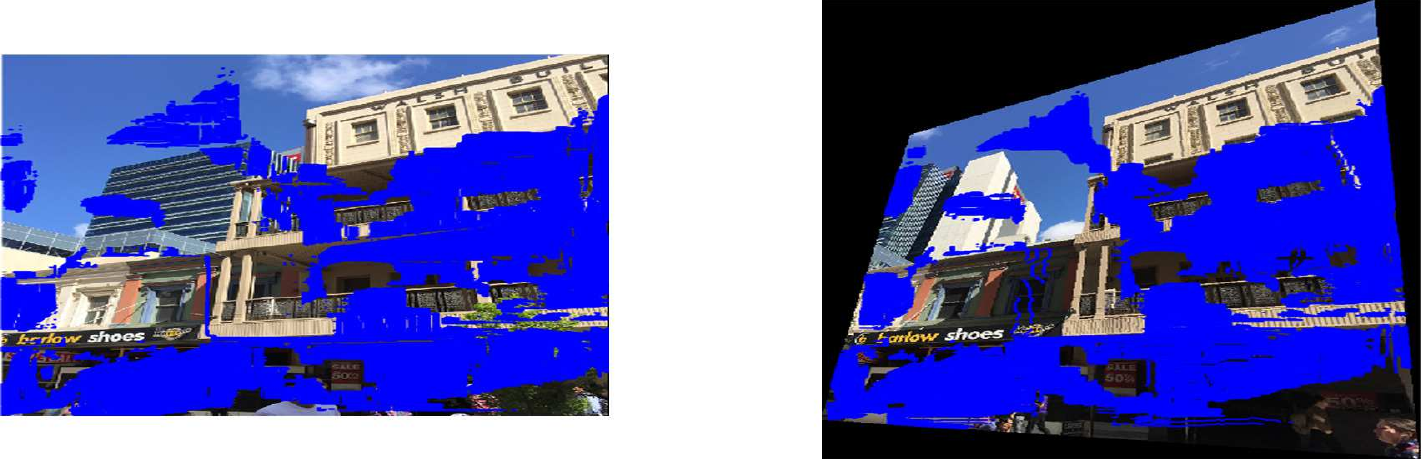}}

\subfigure[Image stitching result using APAP warp~\cite{zaragoza13} estimated from the correspondences in (e).\label{fig3:sf}]
{\includegraphics[width = 0.75\linewidth,height = 0.13\textheight]{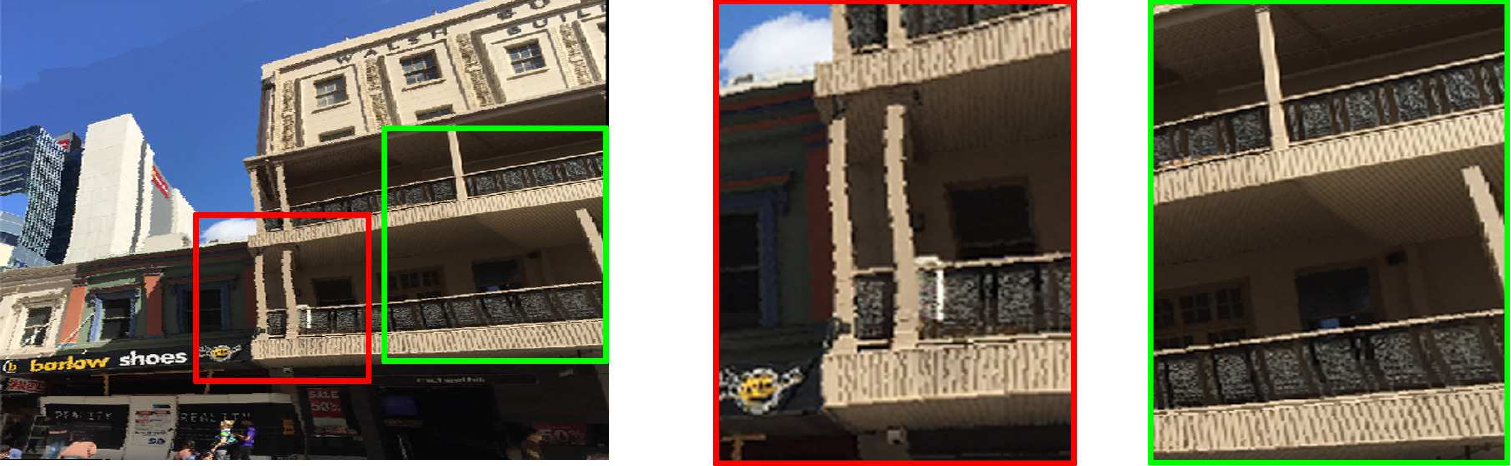}}

\caption{Dense correspondences and stitching results of three flow-based methods on the~\emph{shop front} image pair.}
\label{fig3}
\end{figure}

\begin{figure}
\centering
\subfigure[Pre-warp input images using a homography estimated from SIFT keypoint matches.
\label{fig4:inputs}]
{\includegraphics[width = 0.75\linewidth,height = 0.13\textheight]{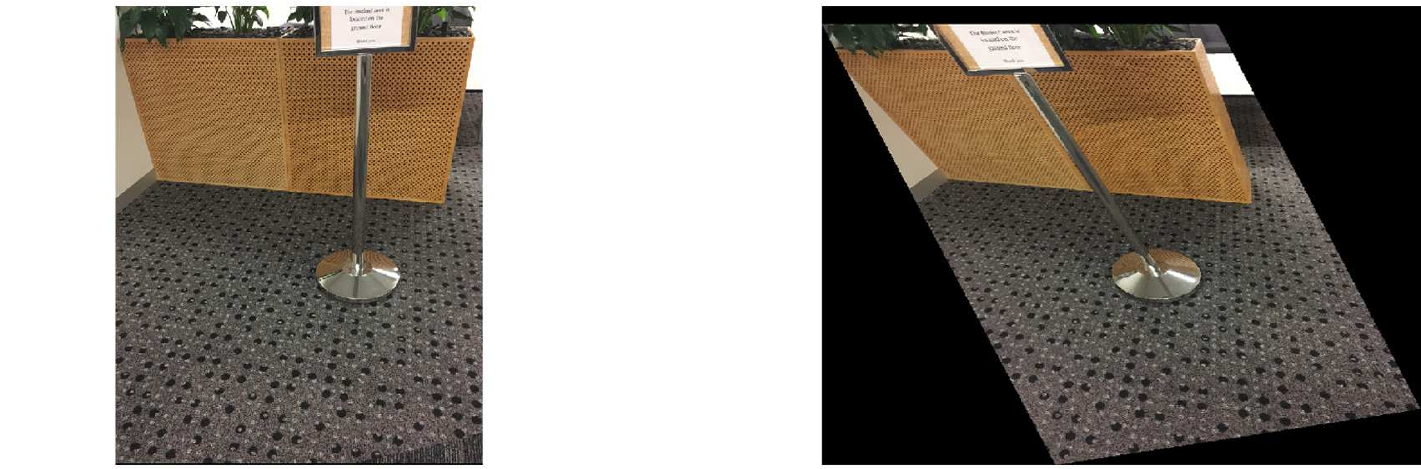}}

\subfigure[Image stitching result using APAP warp~\cite{zaragoza13} estimated from a set of semi-dense correspondences (validated by RANSAC and shown as red points in the final stitched image) produced by Large Displacement Optical Flow~\cite{brox11}.
\label{fig4:tb}]
{\includegraphics[width = 0.75\linewidth,height = 0.13\textheight]{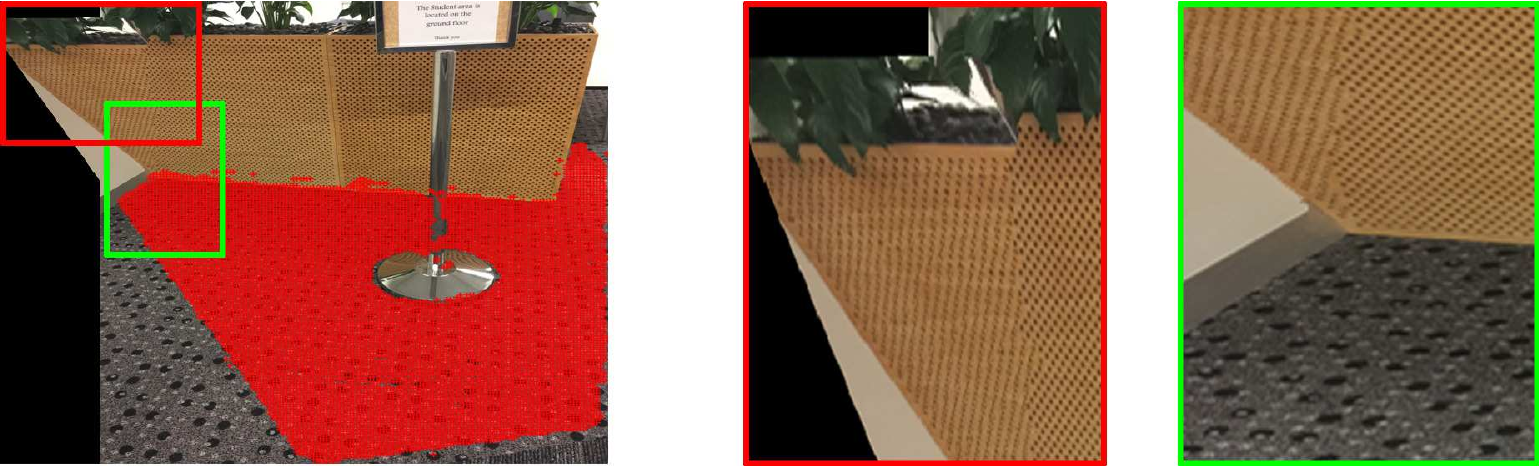}}

\subfigure[The optic flow implementation of~\cite{liuce09} produced 118823 correspondences (after RANSAC validation).
\label{fig4:of_fpts}]
{\includegraphics[width = 0.75\linewidth,height = 0.13\textheight]{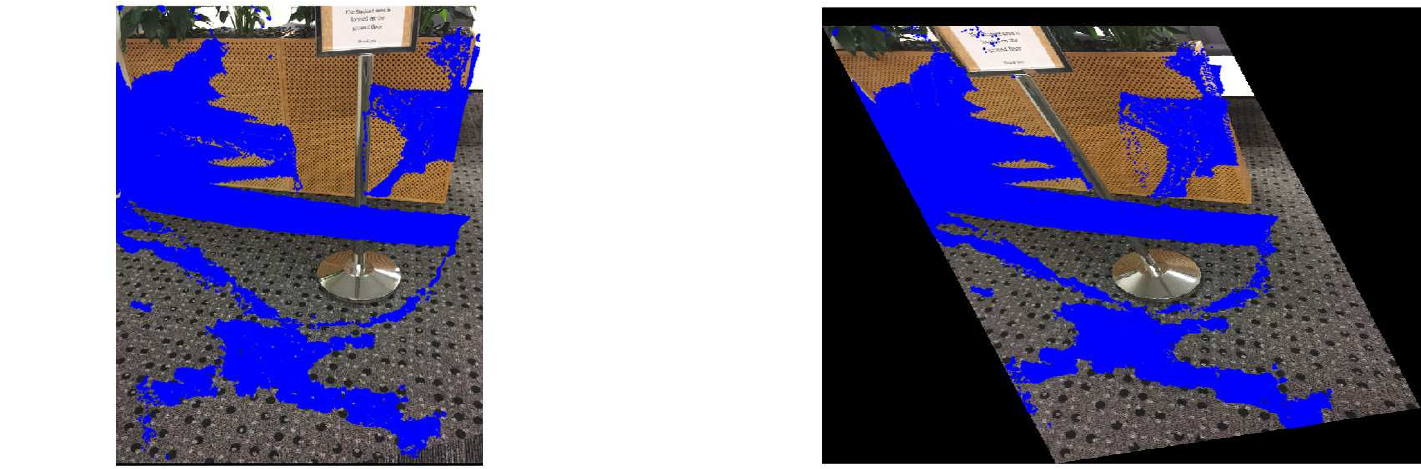}}

\subfigure[Image stitching result using APAP warp~\cite{zaragoza13} estimated from the correspondences in (c).
\label{fig4:of}]
{\includegraphics[width = 0.75\linewidth,height = 0.13\textheight]{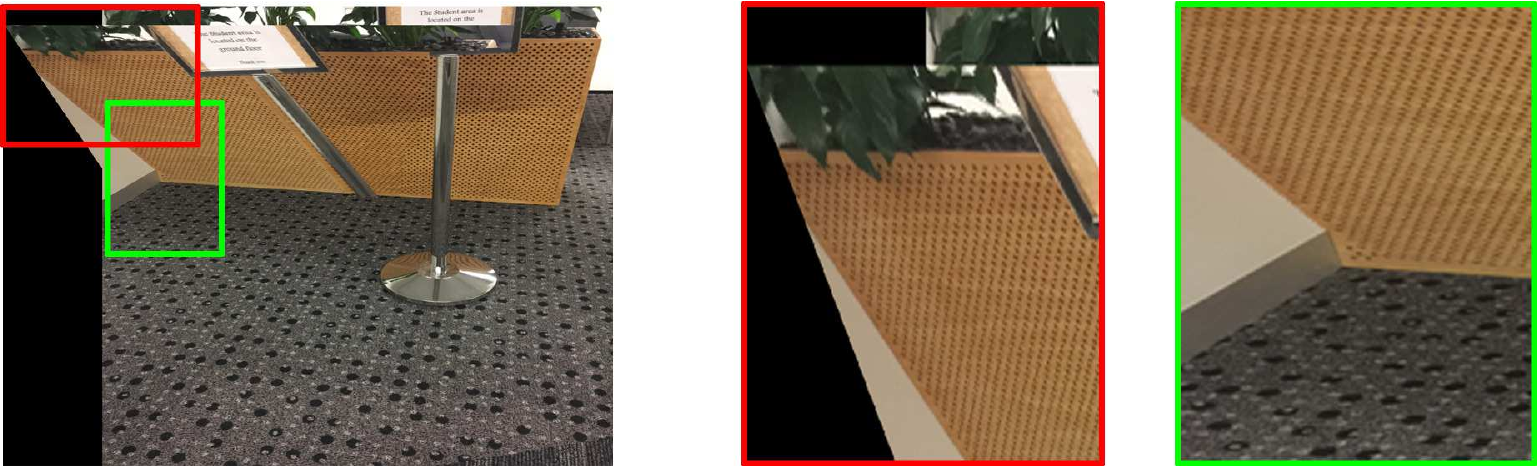}}

\subfigure[SIFT Flow~\cite{liuce11} produced 241276 correspondences (after RANSAC validation).
\label{fig4:sf_fpts}]
{\includegraphics[width = 0.75\linewidth,height = 0.13\textheight]{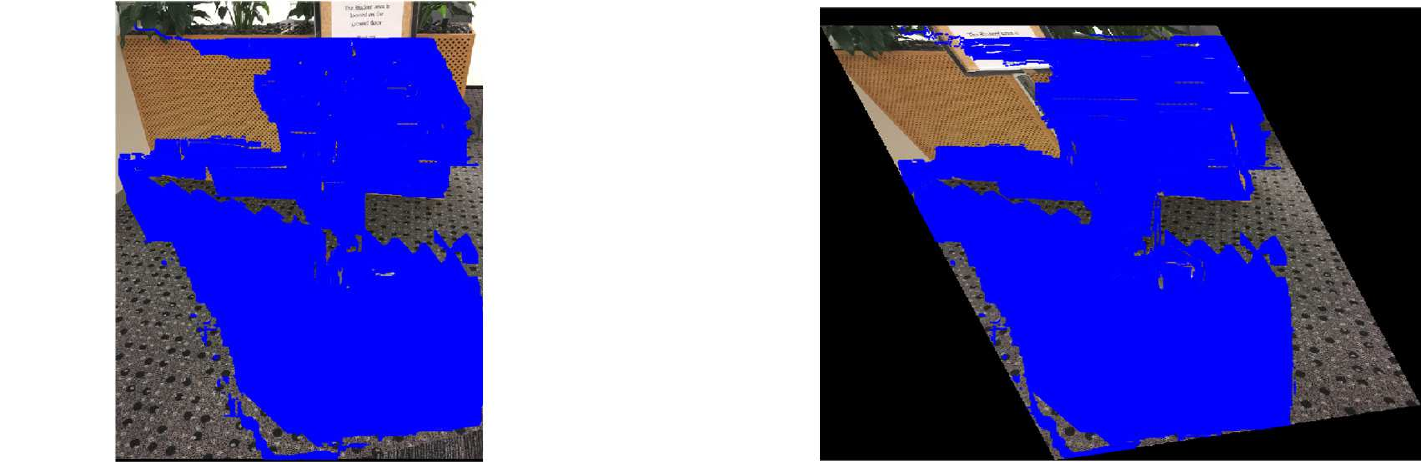}}

\subfigure[Image stitching result using APAP warp~\cite{zaragoza13} estimated from the correspondences in (e).\label{fig4:sf}]
{\includegraphics[width = 0.75\linewidth,height = 0.13\textheight]{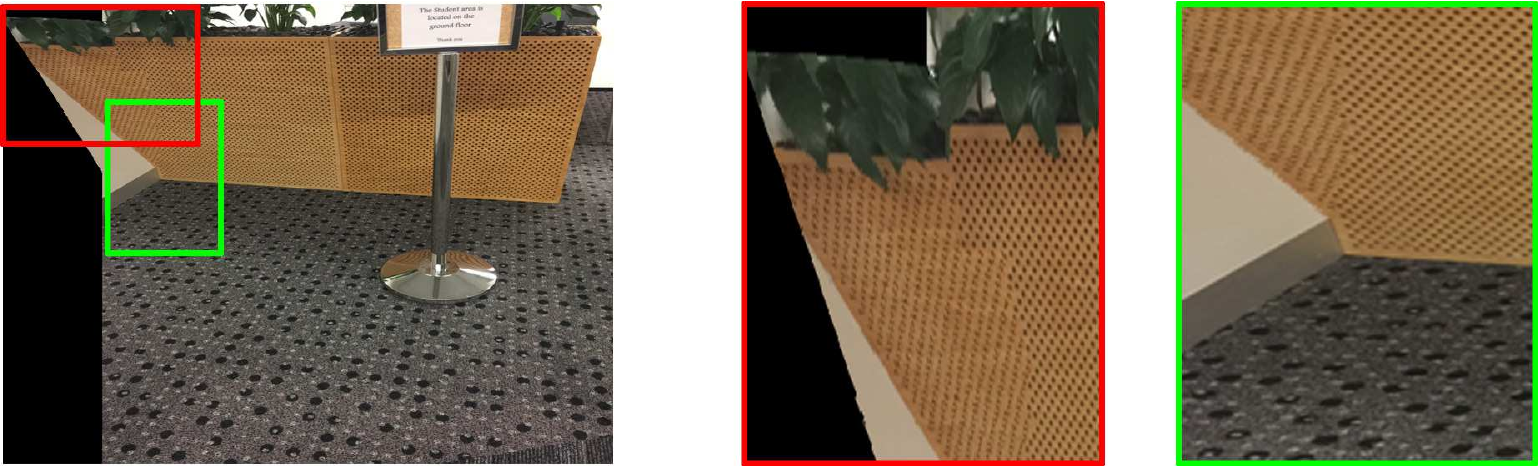}}
\caption{Dense correspondences and stitching results of three flow-based methods on the~\emph{lobby} image pair.}
\label{fig4}
\end{figure}

\hspace{0.5cm}Here, we evaluate the quality of the dense correspondences from flow-based methods for image stitching. The same settings from the main paper (Section 4) were applied here, i.e.,
\begin{enumerate}[noitemsep]
\item One of the images from each pair were pre-warped using a homography estimated from sparse SIFT keypoint matches; see Figs.~\ref{fig1:inputs},~\ref{fig2:inputs},~\ref{fig3:inputs} and~\ref{fig4:inputs}. This served to simplify the data for the flow-based methods.
\item A flow-based method~\cite{brox11,liuce09,liuce11} was invoked to obtain dense correspondences.
\item RANSAC was run with a tight inlier threshold (1 pixel) to validate the correspondences. Typically, even after RANSAC, a very large number of correspondences remained, e.g., on the \emph{truck} datatset, the optic flow implementation of~\cite{liuce09} found $195868$ correspondences, and SIFT Flow~\cite{liuce11} found $403308$ correspondences.
\item The APAP warp~\cite{zaragoza13} was used as a baseline method to stitch each image pair using the dense correspondences.
\item Seam cut pixel selection~\cite{agarwala04} is applied after alignment to remove ghosting and alignment errors.
\end{enumerate}
Figs.~\ref{fig1},~\ref{fig2},~\ref{fig3} and~\ref{fig4} show the results.

\hspace{0.5cm}In Fig.~\ref{fig1}, all three methods produce significant distortions on the chimney and the truck, despite the dense correspondences. In Fig.~\ref{fig2}, dense correspondences were obtained around the pavilion. However, in Fig.~\ref{fig2:tb} the pavilion is seriously distorted, and in Figs.~\ref{fig2:of} and~\ref{fig2:sf} the two separate pillars of the pavilion are merged into one and the tower in the background also appears discontinuous. This points to the inaccuracies in the dense correspondences. In Fig.~\ref{fig3}, the inaccuracies in the dense correspondences are obvious, especially around top left corner of Fig.~\ref{fig3:sf_fpts}. This leads to discontinuities in the balcony and missing pillars.

\hspace{0.5cm}The scene in Fig.~\ref{fig4} contains two apparent planes, and SIFT was able to find good sparse correspondences from only one of them (the floor). Thus, the prewarping result using a homography cannot wholly align the images well. Inevitably, this causes problems for the flow-based methods. Large displacement optical flow~\cite{brox11} found a large amount of correspondences on the ground; however, the region on the wall of the flower bed was not covered. The optical flow implementation of~\cite{liuce09} and SIFT Flow~\cite{liuce11} captured matches on the wall, but still produced stitching results with significant artifacts in Figs.~\ref{fig4:of} and~\ref{fig4:sf}. This is due to the repeated textures on the wall which lead to inaccurate dense correspondences.


\section{Further comparisons on image pairs with significant depth parallax}\label{sec:parallax}

\begin{figure}
\centering
\subfigure[Pre-warp input images using a homography estimated from SIFT keypoint matches.\label{fig5:inputs}]
{\includegraphics[width = 0.80\linewidth]{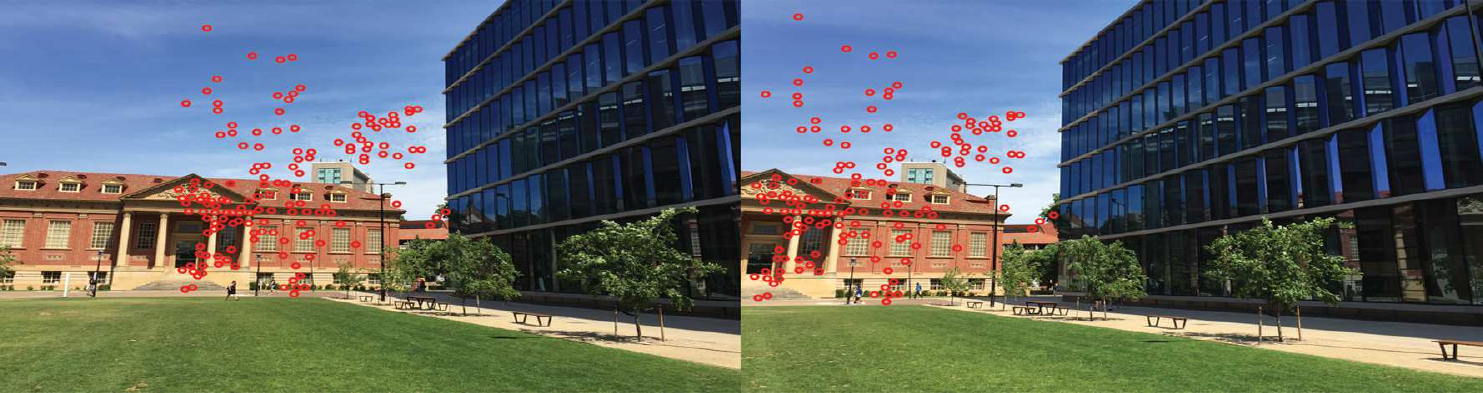}}

\subfigure[Result using parallax-tolerant image stitching.~\cite{zhang14}.\label{fig5:lf_seam}]
{\includegraphics[width = 0.80\linewidth]{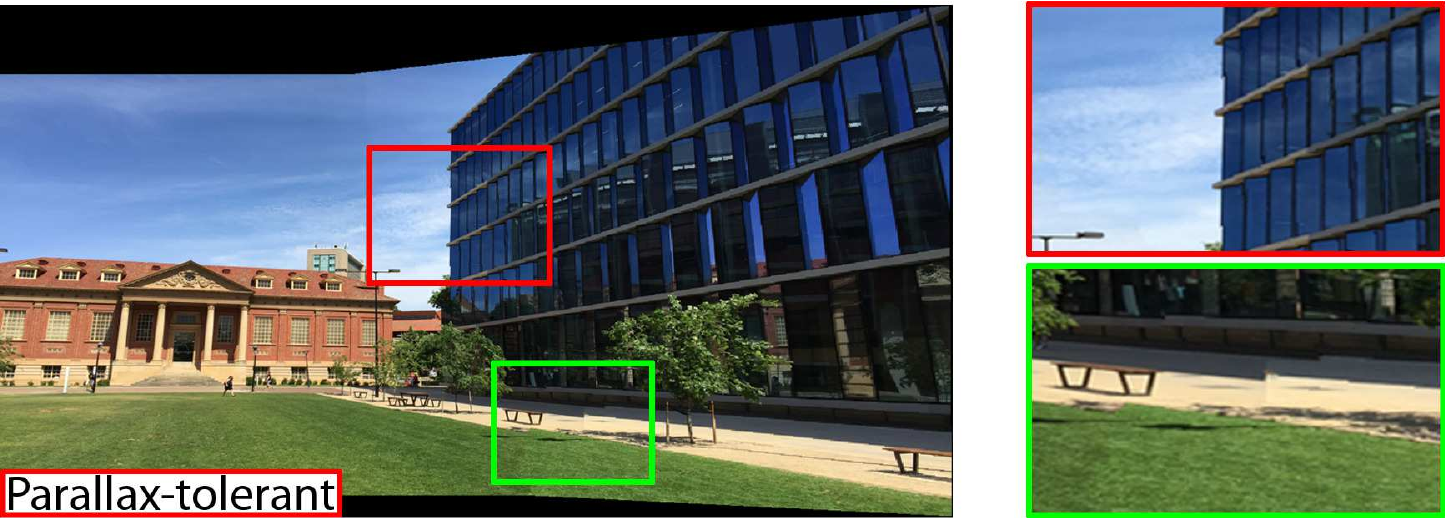}}

\subfigure[Result using APAP warp~\cite{zaragoza13}.\label{fig5:apap_seam}]
{\includegraphics[width = 0.80\linewidth]{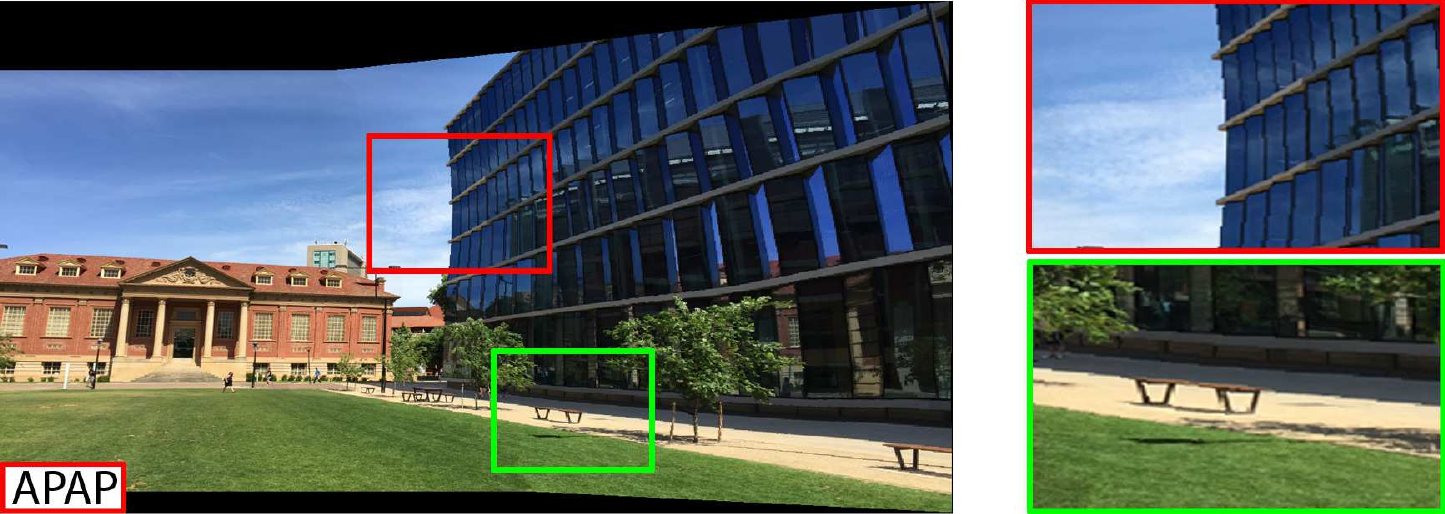}}

\subfigure[Result after adding $25$ new correspondences using the proposed method.\label{fig5:ci_seam}]
{\includegraphics[width = 0.80\linewidth]{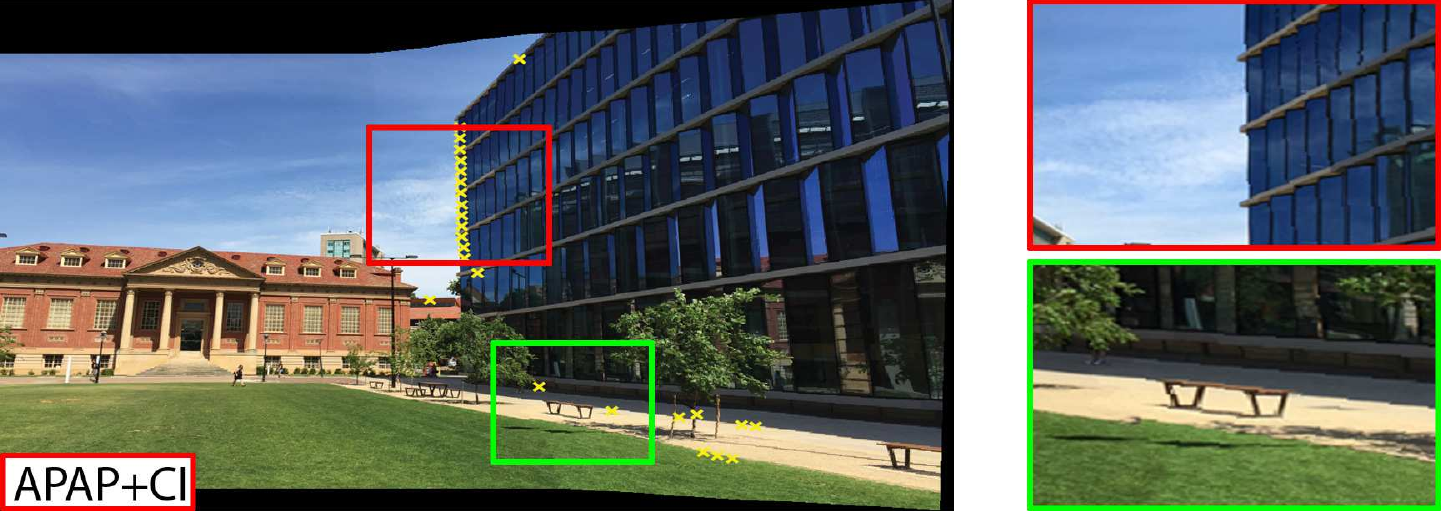}}

\caption{This figure shows a comparison of three methods on the~\emph{lawn} image pair. Inserted correspondences by APAP+CI are shown as yellow points.}
\label{fig5}
\end{figure}

\begin{figure}
\centering
\subfigure[Input images.\label{fig7:inputs}]
{\includegraphics[width = 0.60\linewidth,height=0.22\textheight]{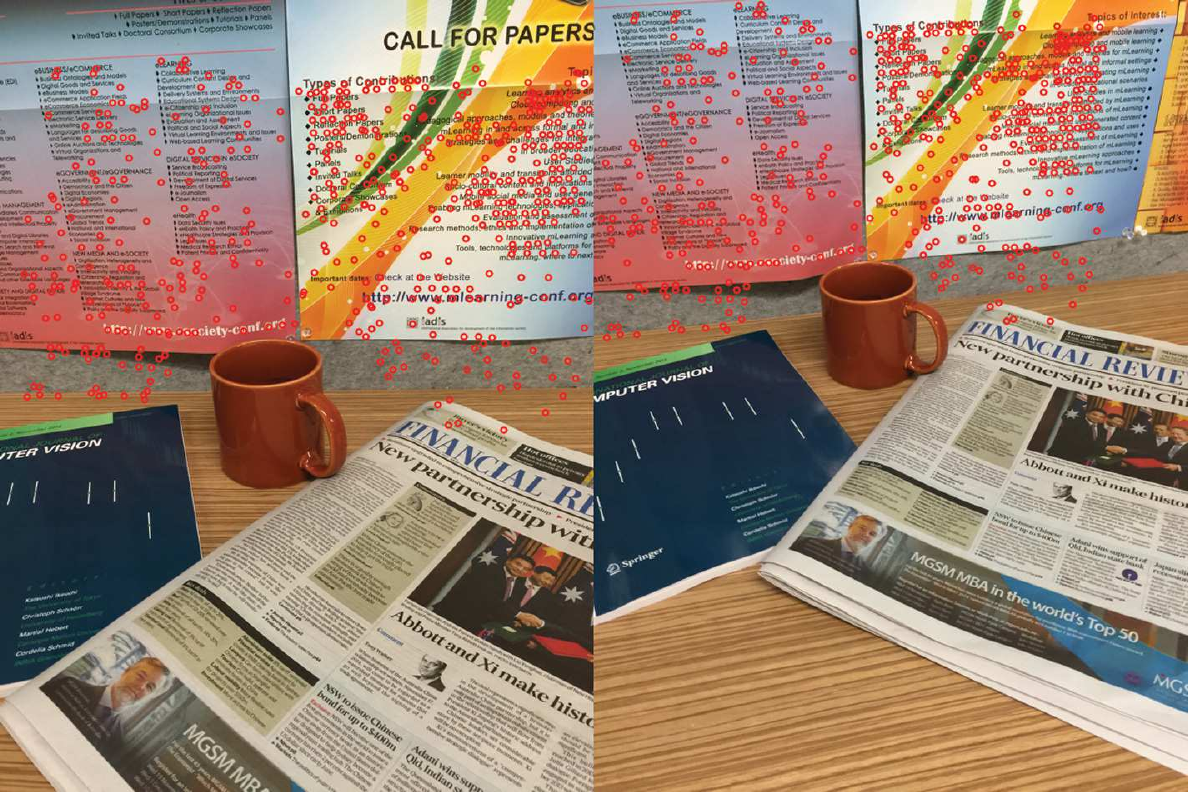}}
\subfigure[Result using parallax-tolerant image stitching~\cite{zhang14}.\label{fig7:lf_seam}]
{\includegraphics[width = 0.60\linewidth,height=0.22\textheight]{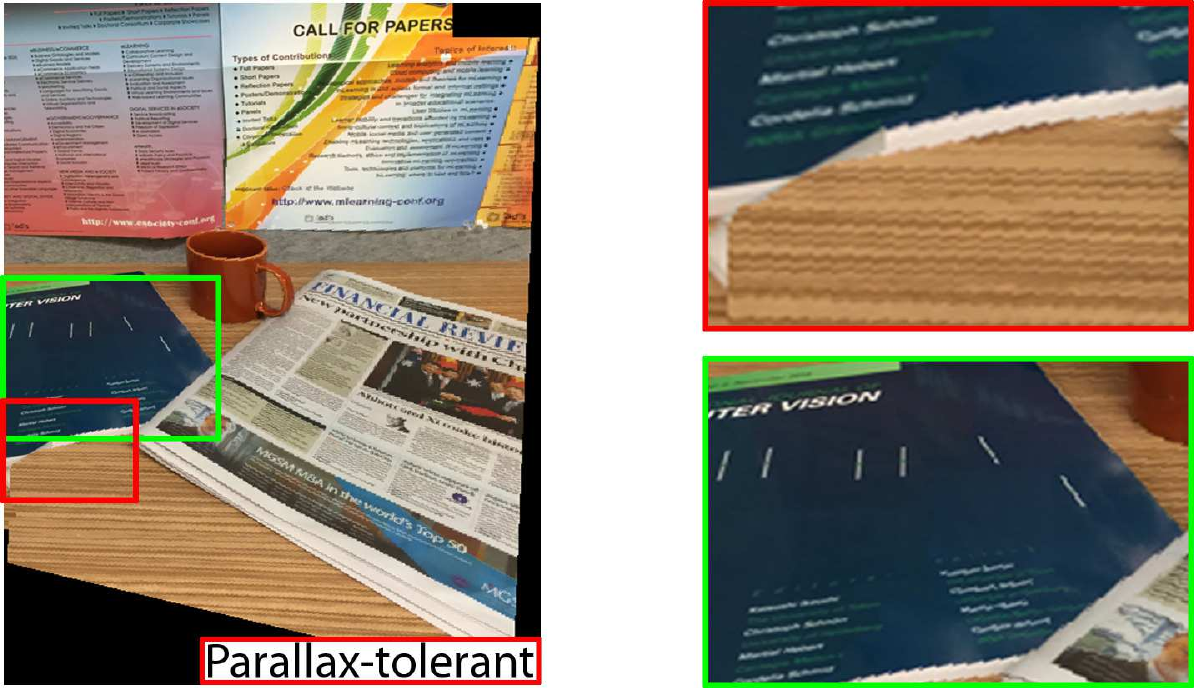}}

\subfigure[Result using APAP warp~\cite{zaragoza13}.\label{fig7:apap_seam}]
{\includegraphics[width = 0.60\linewidth,height=0.22\textheight]{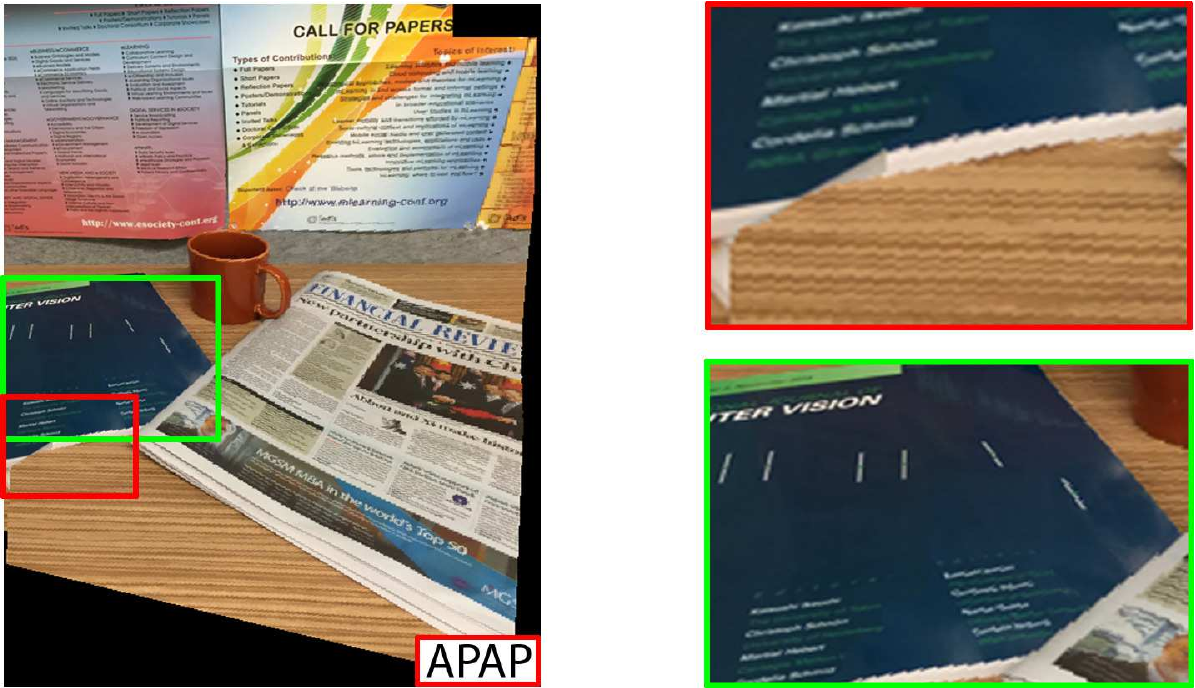}}
\subfigure[Image stitching result after adding $23$ new correspondences using the proposed method.\label{fig7:ci_seam}]
{\includegraphics[width = 0.60\linewidth,height=0.22\textheight]{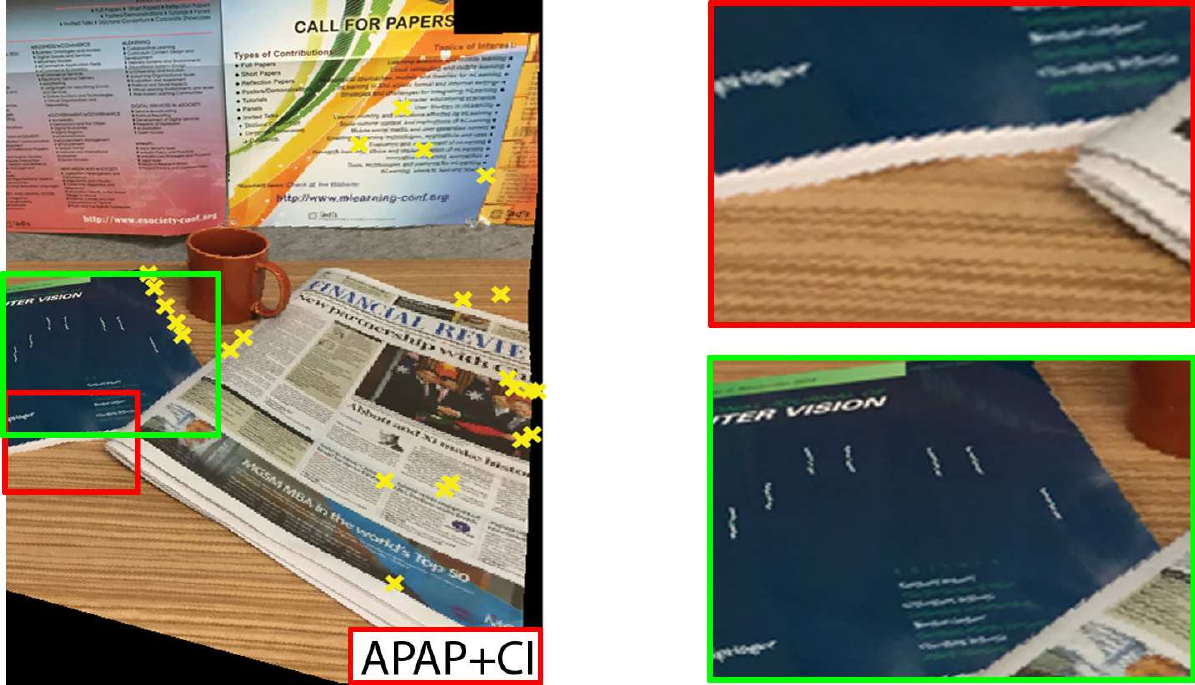}}

\caption{Comparing of three methods on the~\emph{break room} image pair. Inserted correspondences by APAP+CI are shown as yellow points.}
\label{fig7}
\end{figure}

\hspace{0.5cm}In this section, results are shown using images of a scene with significant depth parallax.   We compare our method (abbreviated as APAP+CI) with the state-of-the-art methods, namely the baseline APAP method~\cite{zaragoza13} and the parallax-tolerant image stitching~\cite{zhang14} in Figs.~\ref{fig5} and~\ref{fig7}. Similar to the settings in the main paper (Section 4), here, we apply seam cut pixel selection~\cite{agarwala04} to remove ghosting or alignment errors in the stitched results.

\hspace{0.5cm}In Fig.~\ref{fig5:lf_seam}, parallax-tolerant image stitching introduces notable visual artifacts (green window). In Fig.~\ref{fig5:apap_seam}, APAP produces significant distortion (red window), which is rectified by APAP+CI with the insertion of new correspondences.

\hspace{0.5cm}For the images shown in Fig.~\ref{fig7:inputs}, there are two apparent planes; however most of the feature matches are on one plane only (the wall). Seam cut produced unexpected artifacts in Figs.~\ref{fig7:lf_seam} and~\ref{fig7:apap_seam}. By inserting new correspondences, our method is able to provide a significantly better global alignment.

\section{Comparisons on image pairs without significant parallax}\label{sec:noparallax}

\begin{figure}
\centering
\subfigure[Input images.\label{fig9:inputs}]
{\includegraphics[width = 0.85\linewidth]{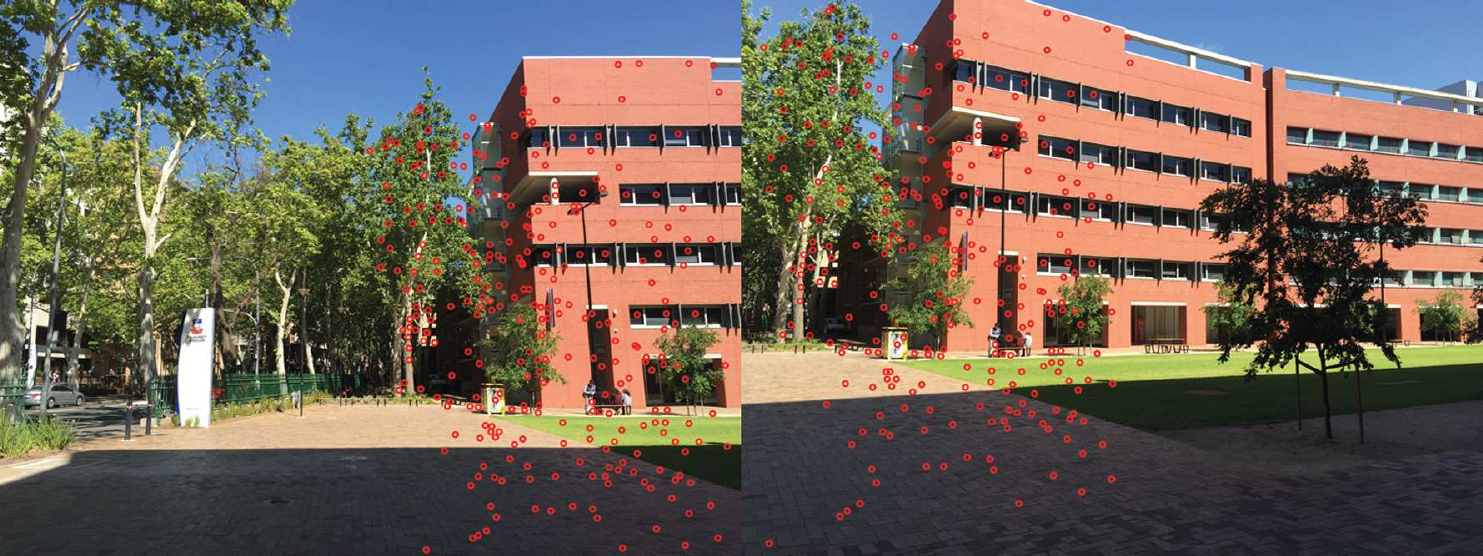}}

\subfigure[Result using a single homography.\label{fig9:lf_seam}]
{\includegraphics[width = 0.85\linewidth,height=0.22\textheight]{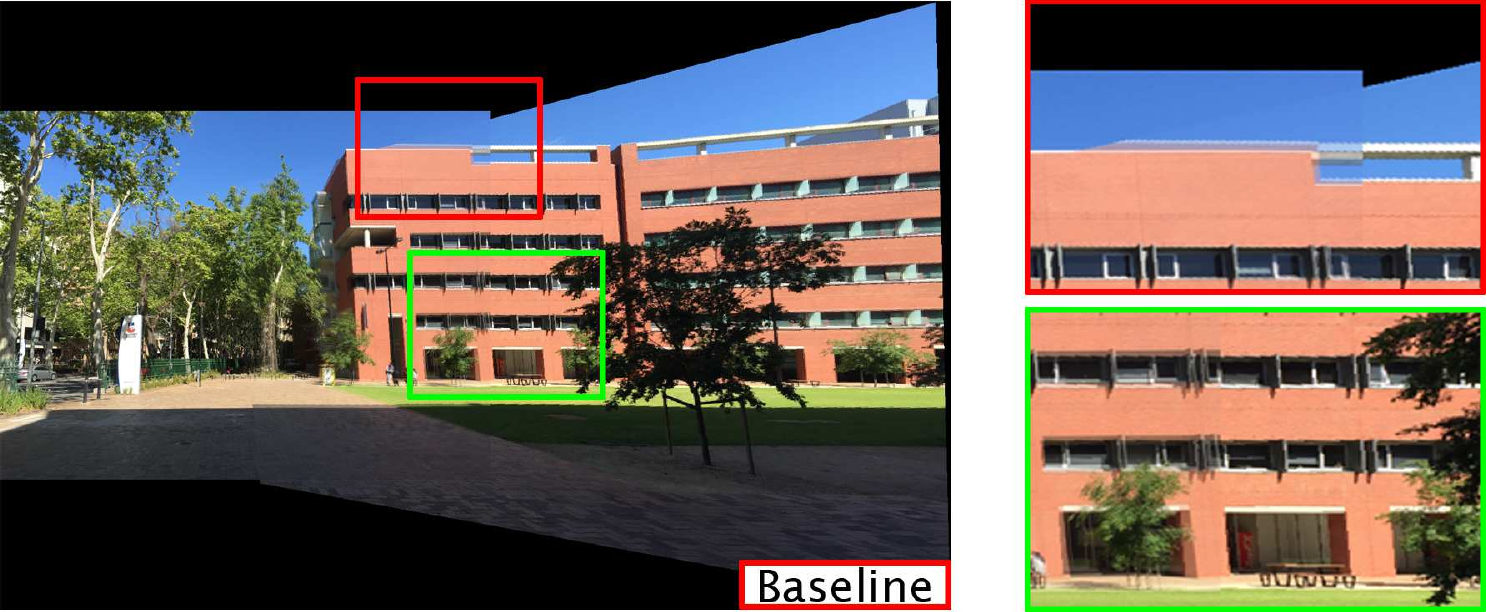}}

\subfigure[Result using APAP warp~\cite{zaragoza13}.\label{fig9:apap_seam}]
{\includegraphics[width = 0.85\linewidth,height=0.22\textheight]{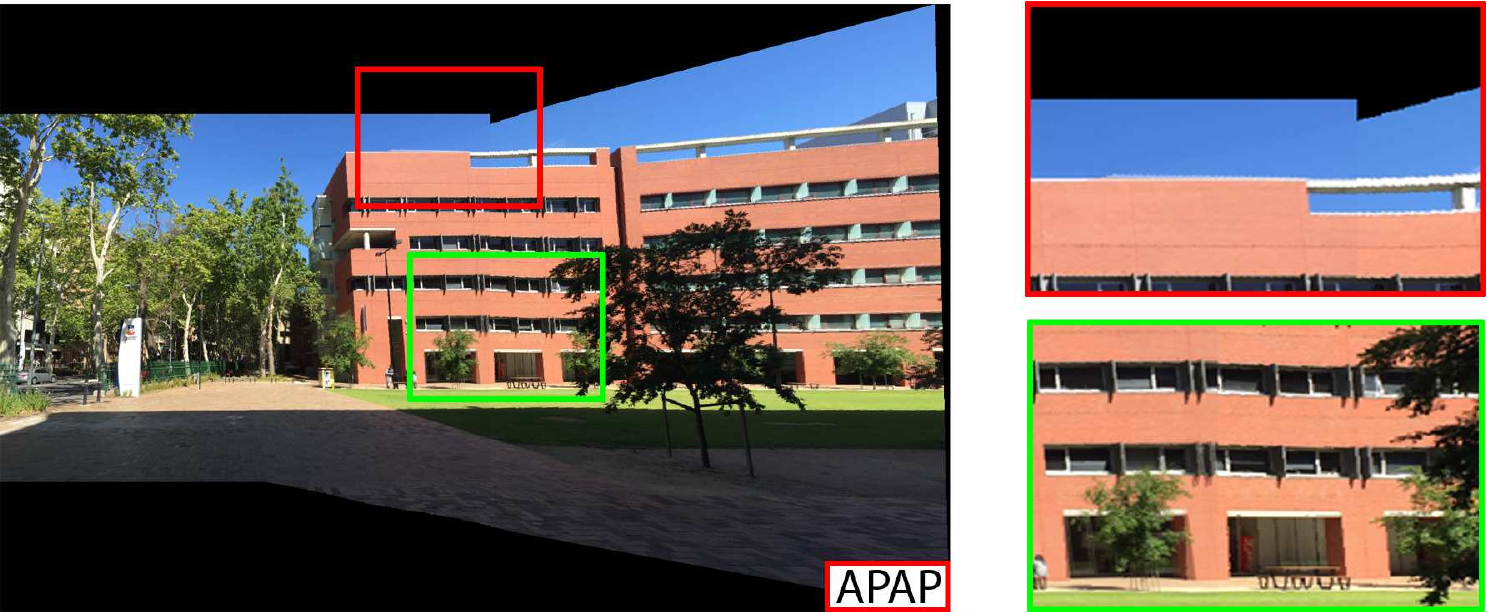}}

\subfigure[Image stitching result after adding $38$ new correspondences using the proposed method.\label{fig9:ci_seam}]
{\includegraphics[width = 0.85\linewidth,height=0.22\textheight]{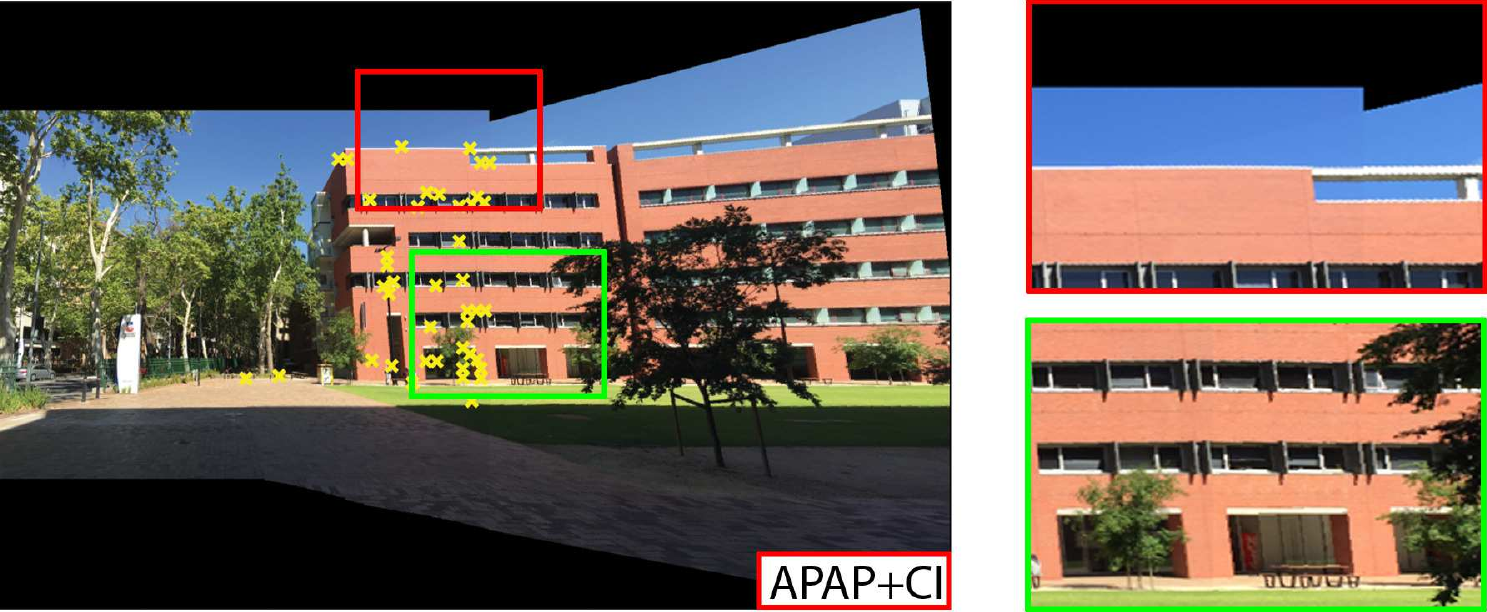}}

\caption{This figure shows a comparison of three methods on the~\emph{building} image pair. Inserted correspondences by APAP+CI are shown as yellow points.}
\label{fig9}
\end{figure}

\begin{figure}
\centering
\subfigure[Input images.\label{fig10:inputs}]
{\includegraphics[width = 0.85\linewidth]{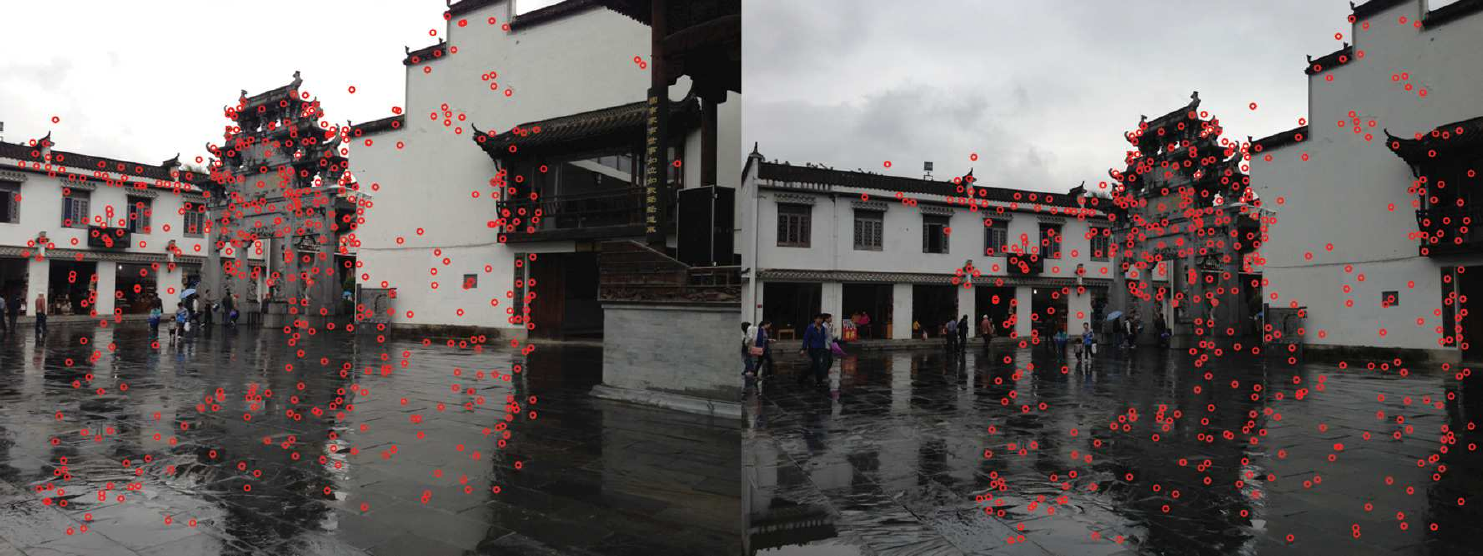}}

\subfigure[Result using a single homography.\label{fig10:lf_seam}]
{\includegraphics[width = 0.85\linewidth,height=0.22\textheight]{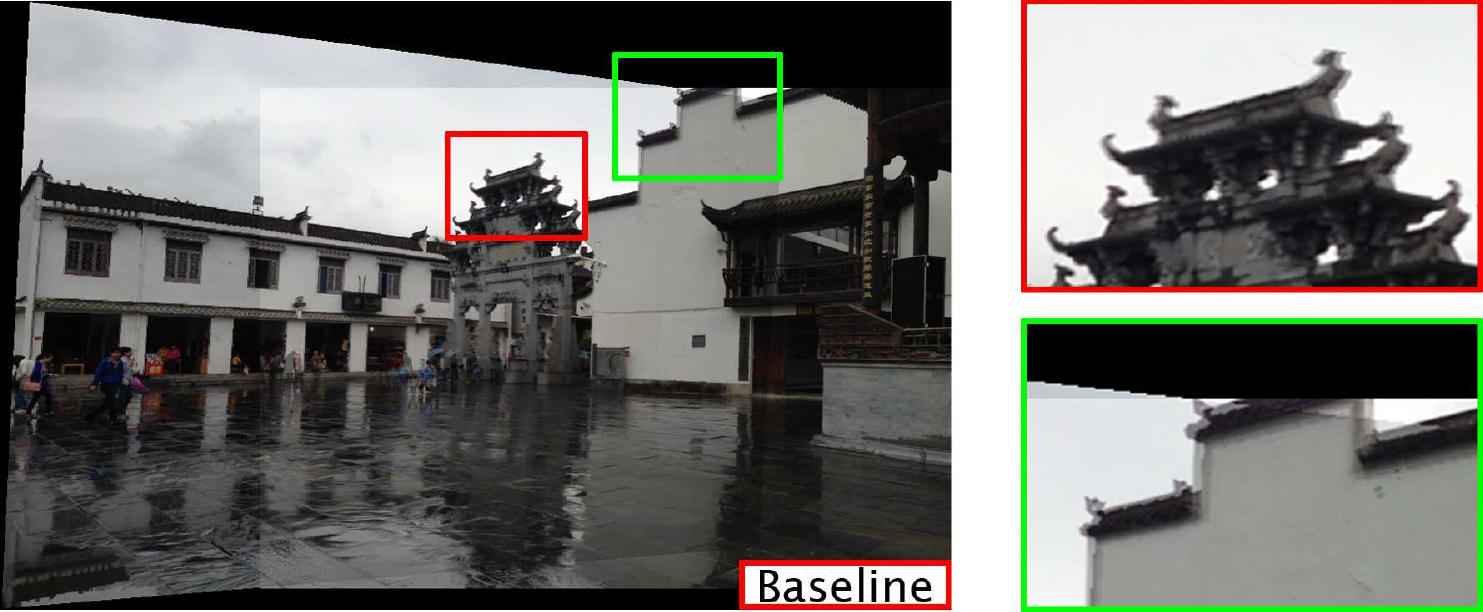}}

\subfigure[Result using APAP warp~\cite{zaragoza13}.\label{fig10:apap_seam}]
{\includegraphics[width = 0.85\linewidth,height=0.22\textheight]{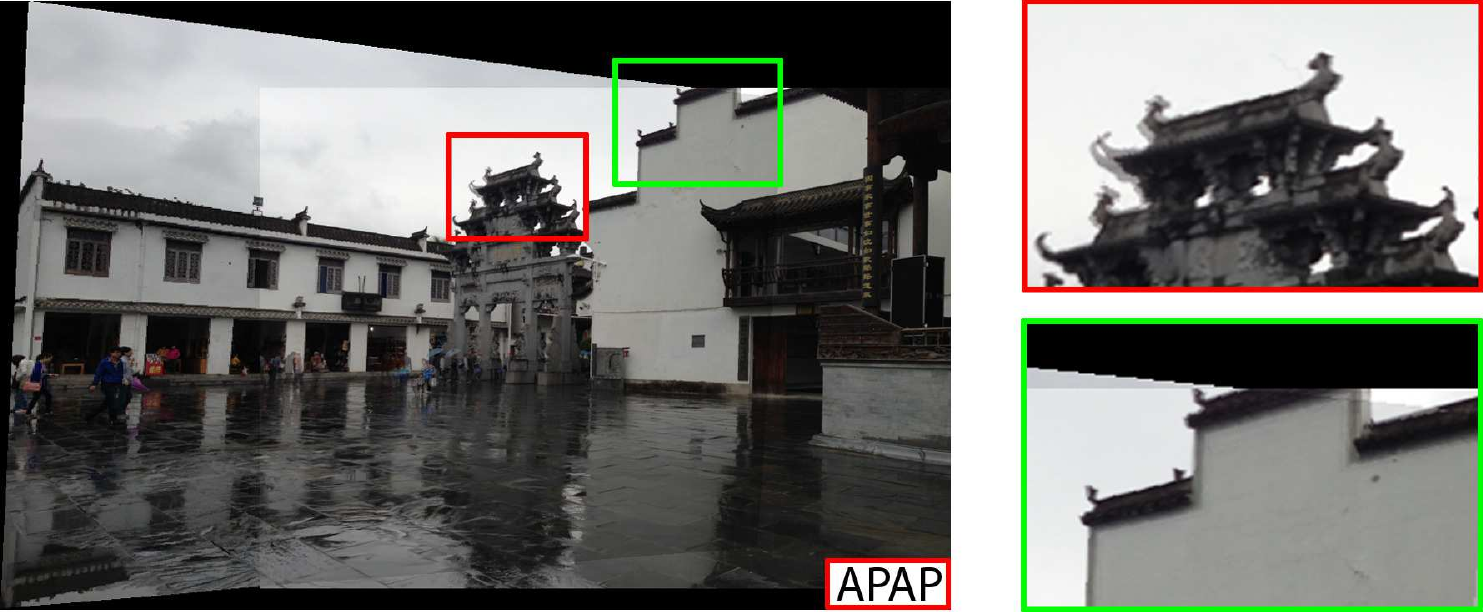}}

\subfigure[Image stitching result after adding $6$ new correspondences using the proposed method.\label{fig10:ci_seam}]
{\includegraphics[width = 0.85\linewidth,height=0.22\textheight]{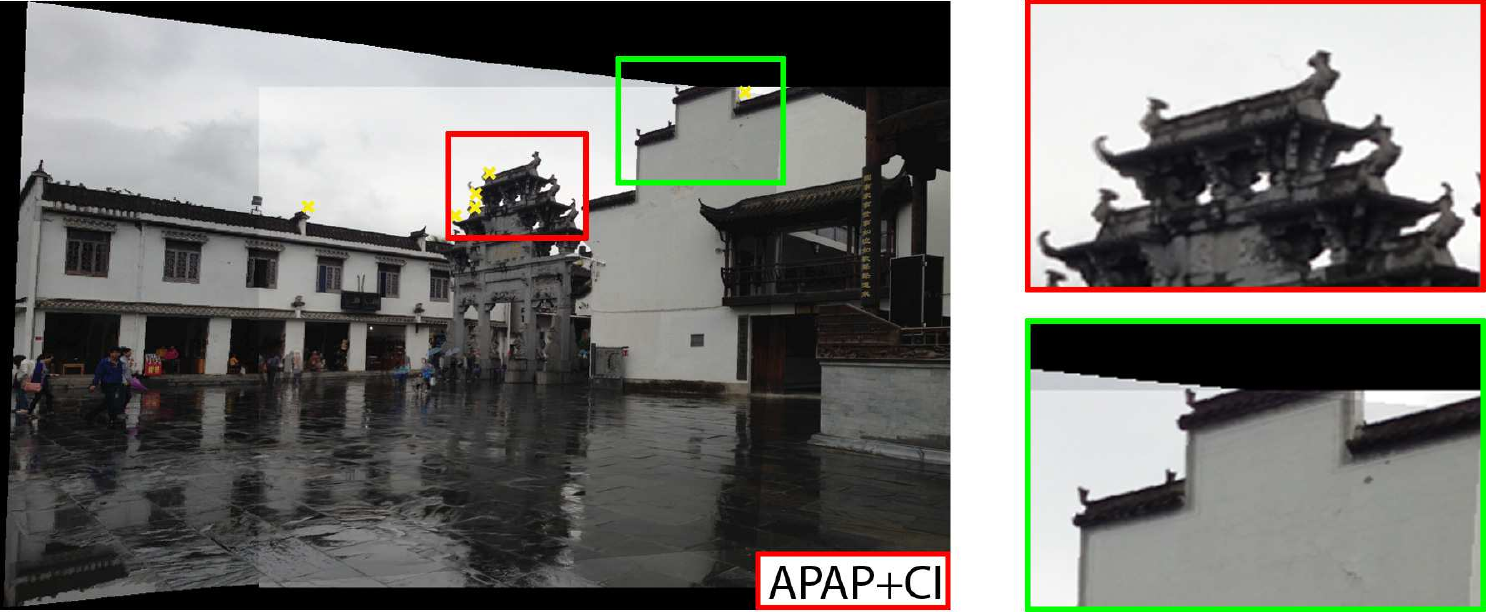}}

\caption{This figure shows a comparison of three methods on the~\emph{arch} image pair. Inserted correspondences by APAP+CI are shown as yellow points.}
\label{fig10}
\end{figure}

\begin{figure}
\centering
\subfigure[Input images.\label{fig11:inputs}]
{\includegraphics[width = 0.7\linewidth]{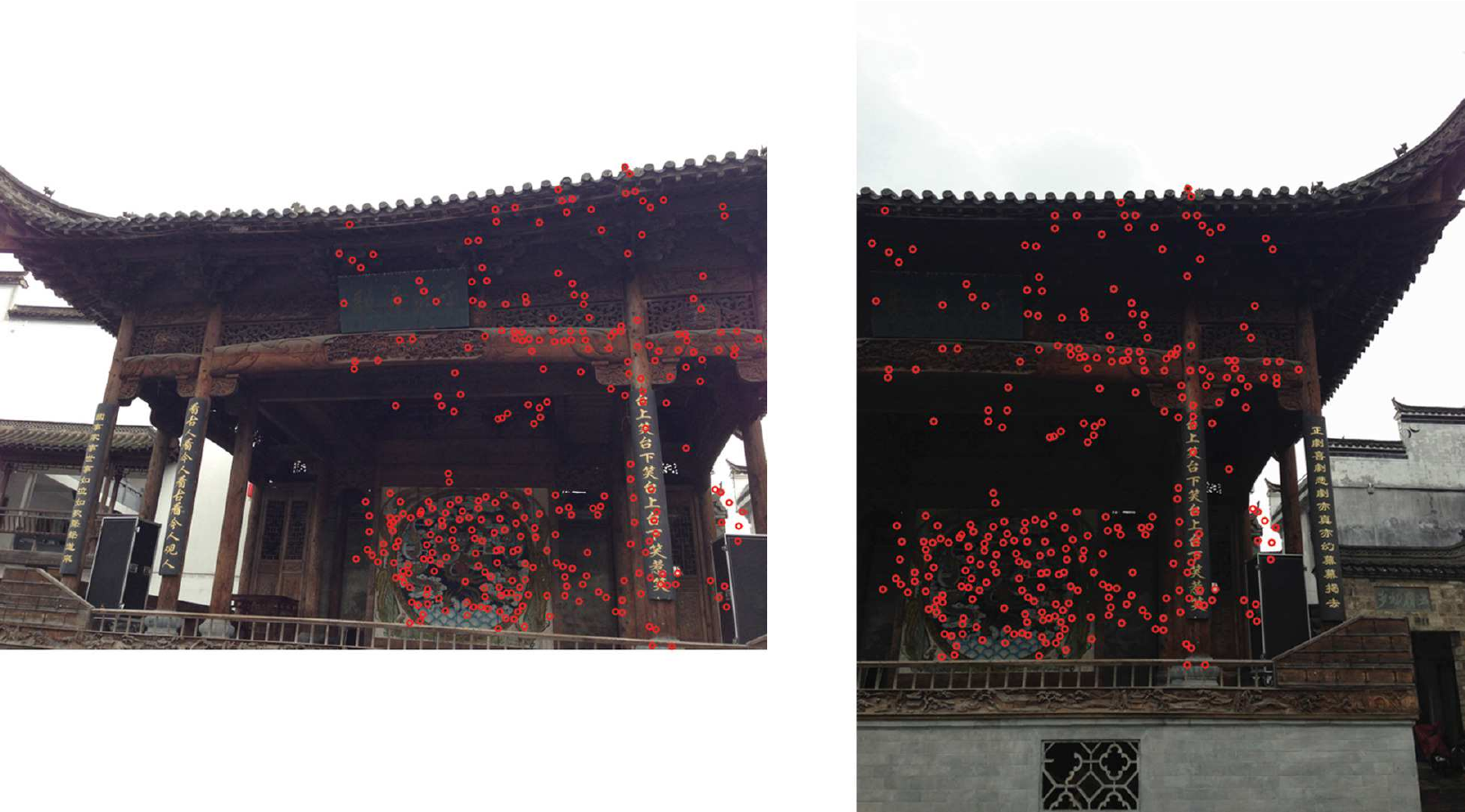}}

\subfigure[Result using single homography.\label{fig11:lf_seam}]
{\includegraphics[width = 0.85\linewidth,height=0.2\textheight]{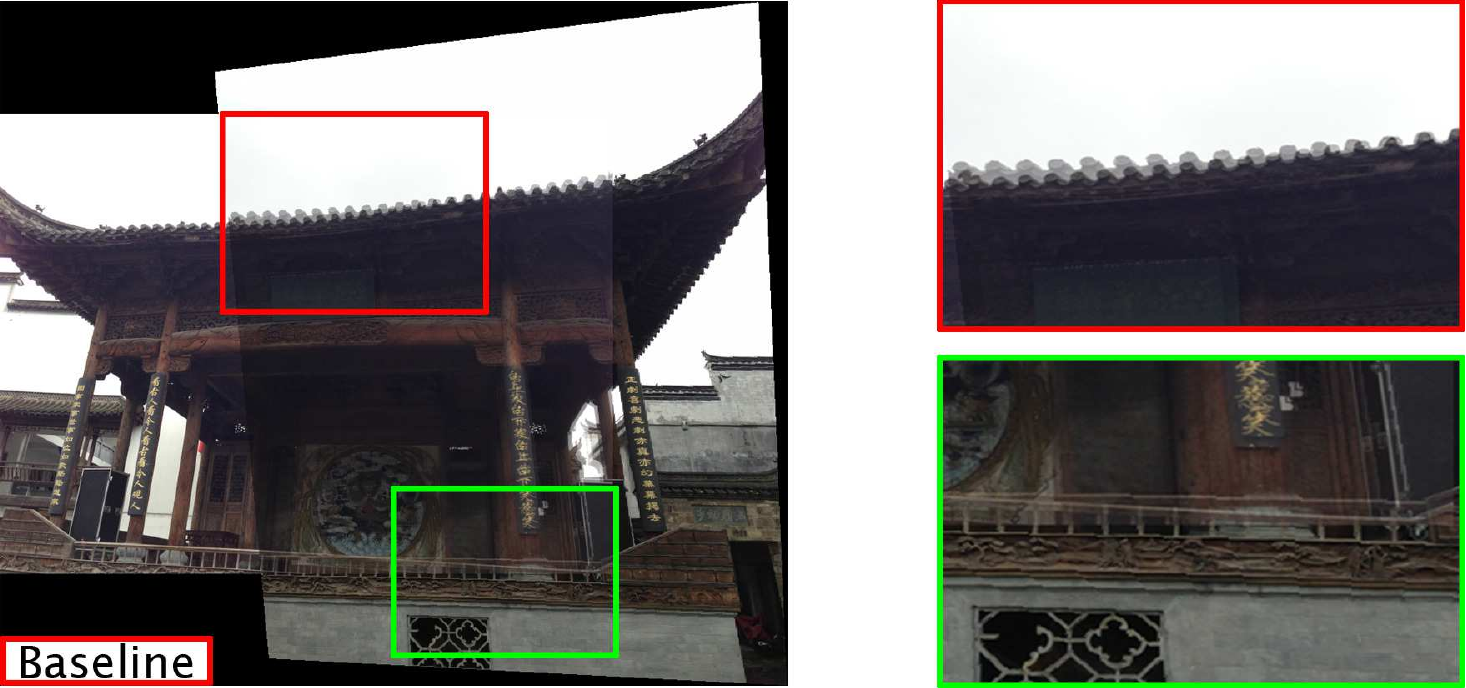}}

\subfigure[Result using APAP warp~\cite{zaragoza13}.\label{fig11:apap_seam}]
{\includegraphics[width = 0.85\linewidth,height=0.2\textheight]{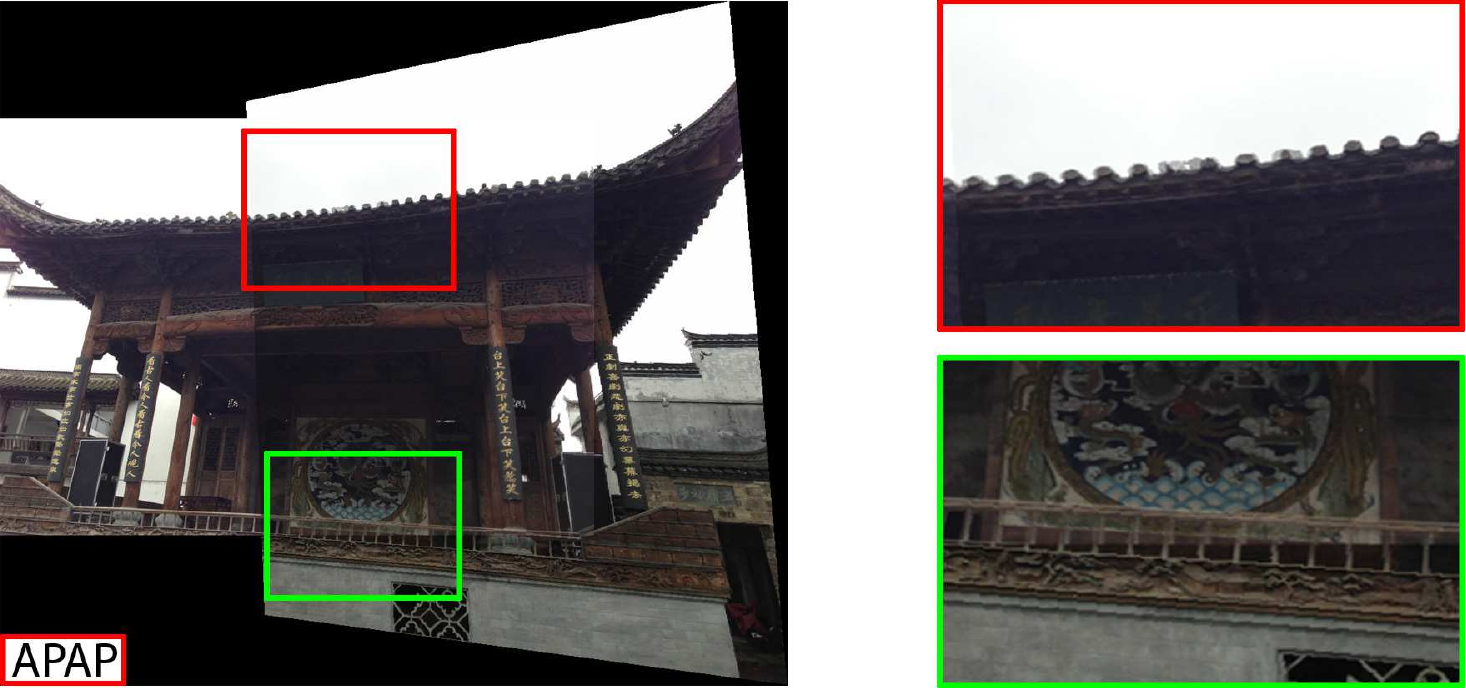}}

\subfigure[Image stitching result after adding $93$ new correspondences using proposed the method.\label{fig11:ci_seam}]
{\includegraphics[width = 0.85\linewidth,height=0.2\textheight]{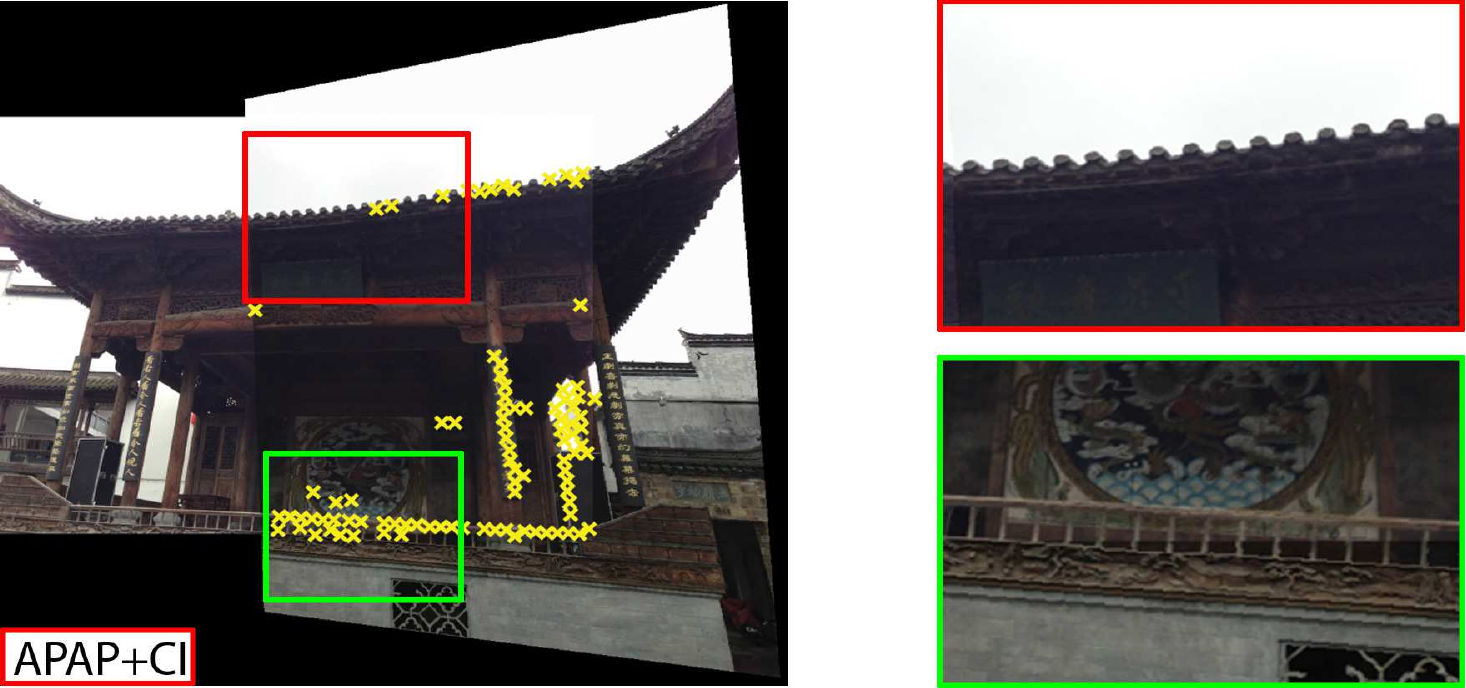}}

\caption{This figure shows a comparison of three methods on the~\emph{stage} image pair. Inserted correspondences by APAP+CI are shown as yellow points.}
\label{fig11}
\end{figure}

\hspace{0.5cm}In this section, we present results on the additional three images pairs that have taken of scenes without significant parallax.   Because parallax is not present, all results are generated without using seam cut blending. Seam cut, however, is an important step in parallax-tolerant image stitching~\cite{zhang14}, so we only compare our method (APAP+CI) with results obtained from a single homography (baseline) and the APAP method. Since seam cut blending is not used, we run our method on the overlapped image region only, which is slightly different from the Algorithm 2 in the main paper.  Figs.~\ref{fig9},~\ref{fig10} and~\ref{fig11} show the results.

\hspace{0.5cm}The single homography model works under the assumption that the images are sufficiently far away or taken with a camera undergoing pure rotational motion.   For these images, this imaging condition is not satisfied and a single homography is not sufficient to align the images.  As shown in Figs.~\ref{fig9:lf_seam},~\ref{fig10:lf_seam} and~\ref{fig11:lf_seam}, warping with a single homography introduces significant ghosting artifacts in the stitching results. APAP warp is able to provide a more accurate alignment, but fail if there are insufficient point matches (e.g. pillar in Fig.~\ref{fig9:apap_seam}, arch in Fig~\ref{fig10:apap_seam}, and railing and eave in Fig~\ref{fig11:apap_seam}). Our method automatically adds new correspondences and rectifies the weakness of the APAP warp producing results that have a better overall alignment.


\end{document}